\pdfoutput=1
\documentclass[11pt]{article}
\usepackage{acl}
\usepackage{times}
\usepackage{latexsym}
\usepackage{microtype}
\usepackage{graphicx}
\usepackage{amsmath}
\usepackage{amsthm}
\usepackage{booktabs}
\usepackage{algorithm}
\usepackage{algorithmic}
\usepackage{subfig}
\usepackage[utf8]{inputenc}
\usepackage{hyperref}
\usepackage{caption}
\usepackage[compact]{titlesec}
\usepackage[normalem]{ulem}
\usepackage{inconsolata}
\DeclareCaptionLabelSeparator{emdash}{\textemdash}
\newcommand{\nb}{\textit{Name Block}}
\newcommand{\db}{\textit{Description Block}}
\newcommand{\eb}{\textit{Example Block}}

\newcommand{\dropexI}{\textit{Drop Examples}}

\newcommand{\droponeI}{\textit{Drop One}}

\newcommand{\dropallI}{\textit{Drop All}}

\newcommand{\anonplusdropexI}{\textit{Anonymize + Drop Examples}}

\newcommand{\anonplusdropallI}{\textit{Anonymize + Drop All}}

\title{A Simple, Yet Effective Approach to Finding Biases in Code Generation}

\author{Spyridon Mouselinos \\
  University of Warsaw \\
  \texttt{s.mouselinos@uw.edu.pl} \\\And
  Mateusz Malinowski \\
  DeepMind  \\
  \texttt{mateuszm@deepmind.com} \\\And
  Henryk Michalewski \\
  Google, University of Warsaw  \\
  \texttt{henrykm@google.com}
  }

\begin{document}
\maketitle

\begin{abstract}
Recently, high-performing code generation systems based on large language models have surfaced. They are trained on massive corpora containing much more natural text than actual executable computer code. This work shows that current code generation systems exhibit undesired biases inherited from their large language model backbones, which can reduce the quality of the generated code under specific circumstances.

To investigate the effect, we propose the "block of influence" concept, which enables a modular decomposition and analysis of the coding challenges. We introduce an automated intervention mechanism reminiscent of adversarial testing that exposes undesired biases through the failure modes of the models under test. Finally, we demonstrate how our framework can be used as a data transformation technique during fine-tuning, acting as a mitigation strategy for these biases.
\end{abstract}
\section{Introduction}
Large language models (LLM) have recently demonstrated their ability to generate code \cite{alphacode,gpt3,wang2021codet5} or solve challenging programming/math tasks on par with human coders \cite{alphacode,lewkowycz2022solving,chowdhery2022palm};  these models are trained with the data-driven paradigm.  On the other hand, an increasing body of work also questions whether the data-driven approach leads to acquiring reasoning skills \cite{piekos2021measuring,zhang2022paradox,mouselinos2022measuring}, showing that if left alone, it might not be sufficient for achieving truly human-level performance on tasks such as logical or visual reasoning.  In many studied cases, models still rely on various hints in their reasoning process. This work extends the results above, i.e., the lack of reasoning capabilities, to the code generation domain. More specifically, we devise a framework that automatically identifies subtle cues a code generation model might exploit. Changes or removal of those cues stands as a reasoning test towards the generational capabilities of the model at hand.

\begin{figure*}[!ht]
\includegraphics[width=0.5\linewidth]{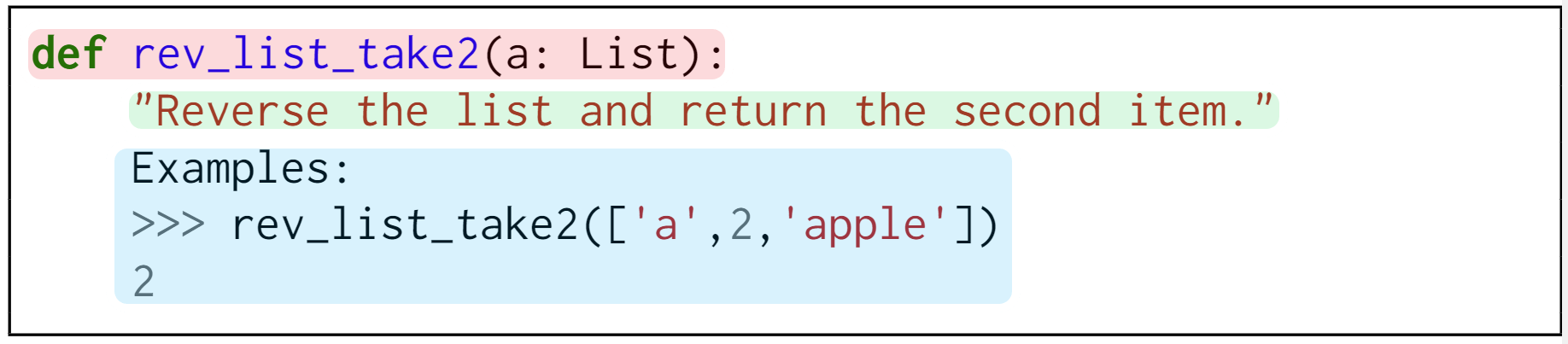}
\includegraphics[width=0.5\linewidth]{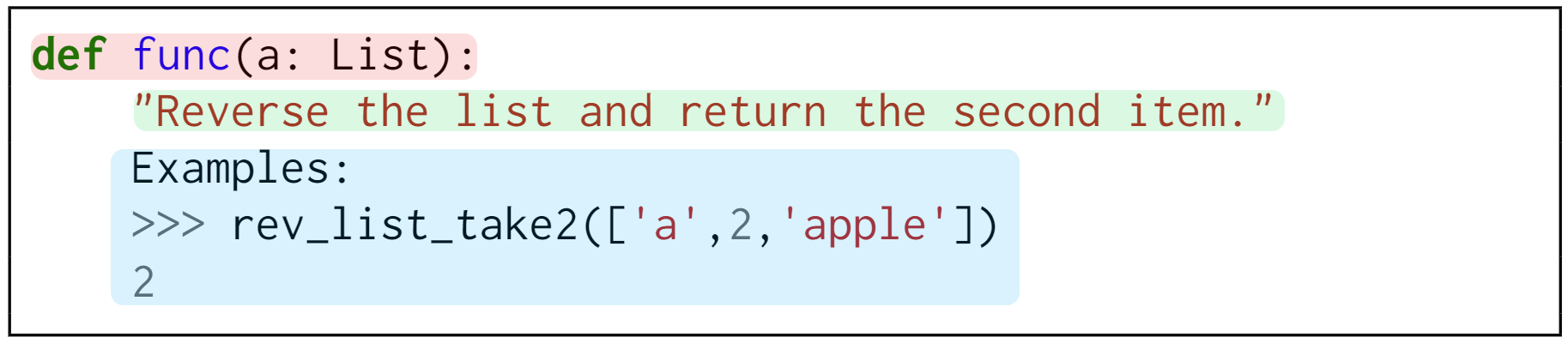}
\caption{\small{\textbf{Left:} The three blocks of influence:  \nb~ in red, \db~ in green and \eb~ in blue. \textbf{Right:} We demonstrate three possible transformations, one for each block: Swap the function name with "func", remove keywords, and remove examples. Transformations can be applied alone or in combinations of two as described in Section \ref{sec:framework}}}%
\label{fig:regions}
\end{figure*}

We presume that the reasoning process of code generation models should remain invariant under changes that still provide enough context or pose little, if any, additional challenge to a human coder. To this end, we propose an automatic and model-agnostic framework that modifies the following: (1) function names, (2) keywords in a problem specification, and (3) examples provided in the problem prompt.
We refer to these parts as Blocks-Of-Influence; see Figure~\ref{fig:regions}. Each block contributes partially to the context needed for correct completion. We show that minor modifications of these blocks are sufficient to ``fool'' LLM-based code generation methods. 

Our results reveal biases such as  keyword preference and memorization effects, which can be identified across multiple models. During our experiments, we ensure that any modifications maintain the global semantics of the coding challenge. This is achieved through a context-aware filtering mechanism that guarantees any information altered or removed still exists and/or can be deducted from the remaining unaltered part.
\paragraph{Contributions.}
The main contributions of our work can be summarized in three points.\newline
\textbf{\textit{First}}, we propose a novel automated framework that identifies possible biases in code generation models. Our framework removes subtle hints, introducing minimal changes such as keyword replacement or partial code-block omission, ultimately acting as an adversarial test. Since the framework operates on a data level, it is agnostic to the model's structure and internal workings. The framework can be easily adjusted to any input format or programming language.
\newline\textbf{\textit{Second}}, we introduce the "\textit{Blocks of Influence}" concept. We suggest that every instance of a typical coding challenge can be analyzed into three parts (blocks). Each part is correlated with a different method of hinting and is used as a target of our transformations. A model's reasoning process is informed by all three blocks, making them perfect analyzing tools for cases of failing code generation.\newline
\textbf{\textit{Third}}, we explore new ways of mitigating biases during code generation. In Section \ref{sec:exp}, we study the effects of adversarial training against our proposed perturbations, and the benefits of including examples with longer descriptions during fine-tuning. Our results show that combining these techniques leads to more accurate code completions.

\section{Related Work}
\label{sec:related_work}
Our approach is inspired by works of various research directions, which we briefly describe here.
\newline\textbf{Solving coding and math challenges.}
The emergent abilities of large language models to generate, summarize and translate textual information, have recently sparked interest in their aptitude for math, logic, and programming challenges. Tasks such as code-completion ~\cite{chen2021evaluating,shin2019program,hendrycksapps2021,alphacode}, code summarization and code translation ~\cite{codeX2021} have been proposed, with models constantly progressing towards near-human performance. Similarly, ~\cite{hendrycksmath2021,deepmindmath2018,aqua2017,mathqa2019} have proposed tests measuring a model's ability to perform math and logic, ranging from school problems to competition-grade challenges.
Impressive results in multiple programming languages have also been achieved by decoder-only works~\cite{gpt3,chen2021evaluating}. 
\citet{fried2022incoder} created the first generative model to perform infilling using a novel masking objective. 
Finally, massive-scale models such as ~\cite{palm,minerva} demonstrated breakthrough capabilities in language, reasoning, and code tasks achieving state-of-the-art performance in multiple domains simultaneously. 
\newline\textbf{Social biases in large language models.}
Trained on ever-increasing amounts of publicly available data, large language models have been studied for adopting social biases commonly found among humans.
 \citet{wallace} show that generative models can be conditioned to produce toxic content, with the use of nonsense, adversarial prefixes. Similarly, \citet{ling} suggest that models might adopt biases and social stereotypes found among their training data and provide ways to apply fairness during generation. Countermeasures have been proposed by \cite{zihao,liu}, claiming that sanitized zero-shot examples contribute to mitigating biases during generation.
\newline\textbf{Probing reasoning through cognitive biases.}  There have been notable attempts to systemize intelligence and reasoning as concepts \cite{legg2008machine,chollet2019measure}, yet a few recent works try to approach reasoning, through the analysis of failure modes, caused by biases in deep learning models. \citet{breaking} suggest that natural language inference systems can be easily fooled with a single hypernym/hyponym swap, exhibiting an bias towards specific word choices. Similarly, \citet{birds} prove that numerical commonsense reasoning in LLMs is heavily biased by adjectives describing the object of interest. Concerns against the current data-driven methods have been expressed by \citet{freq}, pointing out that LLMs are more accurate on mathematical challenges that involve terms significantly more frequently in their pre-training dataset. \citet{piekos2021measuring} claim that LLMs can answer math and logic questions without understanding the rationale behind them, relying blindly on the existence of specific keywords.
We place our work in this line of research, provoking and studying the failures of LLMs under reasoning-heavy coding tasks. Our main goal consists of identifying cognitive bias sources, i.e., words, structures, or co-occurrence patterns, that exist in current LLMs, and lead to systematic failures of rationale.
\newline\textbf{Adversarial methods and Language Processing.} NLP community developed excellent methods to prepare adversarial tasks, including the TextAttack framework \cite{textattack}
and sophisticated techniques to elicit adversarial examples from humans, as in \citet{cqa2}, though our work seems to be the first focused on the disciplined construction of adversarial examples for code.
\section{Benchmarks}
In this section, we describe the datasets used in our experiments. We employed widely used coding challenges HumanEval (HE) and MBPP and a more complex dataset with lengthy descriptions of problems (DMCC). More information about the datasets can be found in the Appendix \ref{sec:modelsanddata}.
\newline\textbf{HumanEval (HE).} This is a human-curated problem-solving  dataset described in ~\citet{chen2021evaluating}. 
It consists of 164 original programming challenges assessing language comprehension, algorithms, and simple mathematics. Each problem is presented as an incomplete function, accompanied by a docstring. The docstring contains the task and a few example cases. For each task, we are provided with a set of unit tests. A task is considered solved when all unit tests are passed. 
\newline\textbf{Mostly Basic Python Problems (MBPP).} Introduced in ~\citet{mbpp}, it contains 974 short Python functions designed to be solved by entry-level programmers. Contrary to HumanEval, each task is given through a text description rather than a docstring. Since there are no input-output examples in the prompt, we generate 3 valid pairs using the code solutions provided.  c
MBPP challenges models to perform tasks of imperative control flow, requiring loops and conditionals.
\newline\textbf{DeepMind Code Contests (DMCC).} Is the highly challenging dataset proposed by \citet{alphacode}. The dataset includes problems from the Codeforces platform \cite{codeforces}, Description2Code \cite{desc2code},
and CodeNet \cite{codenet}. We used challenges written in the Python3 language of the training split for our experiments. DMCC contains long descriptions of the problems and input-output examples of the functions to be completed. 

In this work, DMCC is used for its long context properties during experiments of augmented fine-tuning (Table \ref{long_context}). Models presented in our work achieve zero or near-zero scores on it; hence it is excluded from our perturbation analysis, with HumanEval and MBPP being more suitable targets.

\section{Evaluation}
\textbf{Models.} In our experimental setup, we test five models representing different approaches to code generation. CodeParrot~\cite{codeparrot} comes with an open-source dataset and can be easily used for fine-tuning experiments due to its size. Its smaller variant (110M) achieves competitive results to other open-source LLMs at larger parameter budgets. By exploring its dataset, we tested our hypothesis that function names act as biases during code generation. Models can be heavily inspired by similarly named snippets in their training set and resort to copying whole or parts of the solution instead of performing reasoning. (See Appendix A.9)
We also test the Incoder~\cite{fried2022incoder} model, which is trained under a novel bi-directional causal objective, being able to handle context more efficiently than its causal counterparts. Against our initial hypothesis, our methods cause significant performance drops despite the model's enhanced context-understanding capabilities (Table \ref{qual_table1}). The Bloom model ~\cite{bloom2022} exhibits emergent abilities in multiple domains by training on massive multilingual and multi-purpose content. Despite not being a code generation model, it performs equally well with code-specific models in the same parameter budget. Theoretically, bias effects can be reduced when a model is exposed to diverse training examples. Our experiments reveal that this is still not the case under our setup, and post-training solutions are explored. CodeGen~\cite{CodeGen} is a high-performing model trained in natural language understanding and code. We test its Mono variant, further fine-tuned on the Python language. Finally, we have the powerful Codex model, which can tackle most of the proposed coding challenges in the HumanEval and MBPP datasets. A list of the tested models, as well as KeyBert~\cite{grootendorst2020keybert} that is used in our framework, can be found in Table \ref{model_table}.
\begin{table}[!ht]
\centering
\resizebox{0.75\columnwidth}{!}{
\begin{tabular}{lc}
Model Name &  Sizes Used \\
\midrule
KeyBert \cite{grootendorst2020keybert} &   2M\\
Codeparrot \cite{codeparrot}  &   110M / 350M*/ 1.5B    \\
InCoder \cite{fried2022incoder}  &  1.6B / 6B  \\
CodeGen \cite{CodeGen}  &  350M / 6B  \\
Bloom \cite{bloom2022}  &   560M* / 1.7B / 176B$^\dag$  \\
Codex (v1 / v2) \cite{chen2021evaluating}  &   $\sim$175B$^\dag$ (Estimated)
\\
\bottomrule
\end{tabular}
}
\caption{\small{Models used: (*) refers to fine-tuned and (\dag) to API.}}\label{model_table}
\end{table}
\newline\textbf{Performance metrics.} 
We evaluate the functional correctness of the generated programs with the pass@k metric, introduced in \citet{spoc}. This metric serves as an estimator of a model's generative capabilities under a specific budget. In \citet{chen2021evaluating}, authors propose an updated unbiased version that we adopt throughout the rest of this work. To avoid any confusion, we calculate pass@k at exactly k attempts.  The average of ten runs with different seeds is presented for all experiments in Table \ref{qual_table1}. We use sampling temperatures of 0.2 / 0.8 for pass@1 / pass@100, which are the optimal values across the tested models.
\section{Method}
\subsection{Blocks of Influence}\label{subsec4_1}
\begin{figure}[!ht]
%\minipage{\columnwidth}
% \begin{minted}
% [frame=single,
% framesep=1mm, 
% fontsize=\tiny,
% breaklines,
% highlightlines={9}
% ]
% {python}
% from typing import List
% def has_close_elements(numbers: List, threshold: float):
%     "Check if in the given list of numbers, are any two numbers closer to each other than given threshold."
%     Examples:
%     >>> has_close_elements([1.0, 2.0, 3.0], 0.5)
%     False
%     >>> has_close_elements([1.0,2.0,3.0,4.0,5.0,2.0], 0.3)
%     True
%     return any(abs(x-y) < threshold for x,y in zip(numbers, numbers[1:]))
% \end{minted}
%\endminipage \hfill
%\minipage{\columnwidth}
% \begin{minted}
% [frame=single,
% framesep=1mm, 
% fontsize=\tiny,
% breaklines,
% highlightlines={4,5,6,7},
% highlightcolor={pink}
% ]
% {python}
% from typing import List
% def has_close_elements(numbers: List, threshold: float):
%     "Check if in the given list of numbers, are any two numbers closer to each other than given threshold."
%     for i in range(len(numbers) -1):
%         if abs(numbers[i] - numbers[i+1]) < threshold:
%             return True
%         return False
% \end{minted}
%\endminipage
\includegraphics[width=\linewidth]{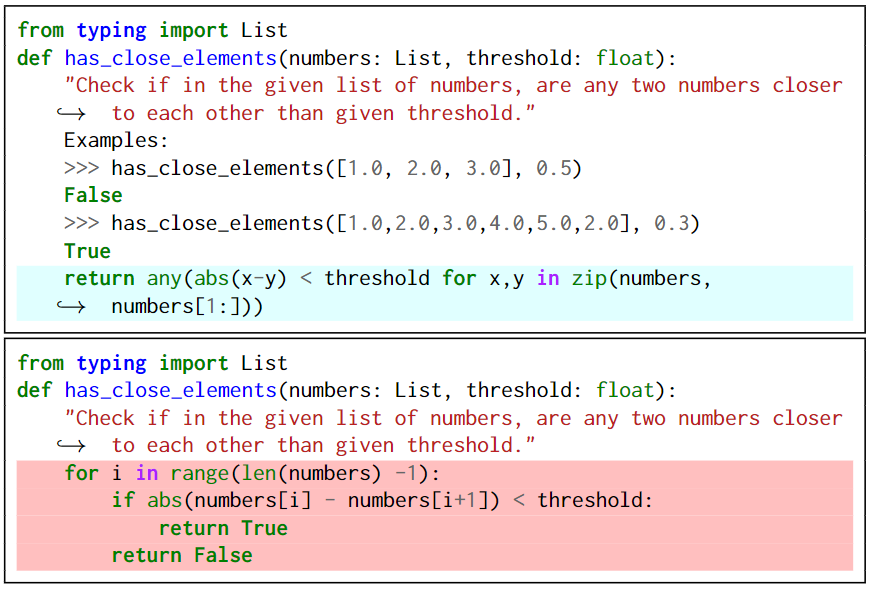}
\caption{\footnotesize{Example removal reveals poor reasoning (\textit{Example drop} / Codex-v1): The model initially exhibits signs of task comprehension (top), generating a correct solution. Removing the examples, however, reveals a lack of proper reasoning; Although the model still understands that it has to compare numbers, it resorts to a naive sequential check instead of comparing each available pair (bottom).}}
\label{fig:temp_2}
\end{figure}

Our method treats each coding challenge as a combination of three distinct but complementary blocks rather than a single, homogeneous input. We refer to them as \textit{Blocks of Influence} and correlate each with a different source of bias during code generation. Taking as an example Figure \ref{fig:regions}, we challenge the model to complete a function that reverses a list and then returns its second item. 
\newline\textbf{Name Block.} The first block of influence, marked in red, informs the model about the function name and the names and types of the input arguments. Let us assume that initially, a model generates correct solutions to a problem. However, the model fails when we rename the function name to something unrelated to the task, e.g., \textit{``fun``}. This failure mode indicates that neither the problem description was understood nor the model could extract a reasoning pattern from the given usage examples. We associate such cases with memorization effects, where the model relies heavily on the function name, replicating snippets from its training dataset with the same or similar names.
\newline\textbf{Description Block.} The problem description stands as the second block, marked in green. Here the model is expected to form a solution by utilizing its natural language understanding capabilities. We observe that removing specific keywords from the problem description can lead to catastrophic results in model performance. It is vital that removing these keywords must not degrade the description semantics, and any information lost should be recoverable from the rest of the context. For example, in Figure \ref{fig:regions}, the removal of the word pair "the list" creates a description that is still well understandable by a human coder. We challenge the model to deduct the missing context from the word "list" in the function name and the input list type in the example given. The inability to recover the missing context is associated with an inherent preference bias, where the model relies on superficial lexical clues or frequently co-occurring terms seen during training rather than the given context to "mentally" fill any gaps.
\newline\textbf{Example Block.} 
As the final block, we consider the examples after the problem description. They act as demonstrations, guiding the model to specific reasoning patterns. Let us consider a scenario where models cannot generate correct code when examples are absent. 
Arguably, more than the task and given inputs alone were needed for the model to form a proper problem understanding. In this failure mode, the provided examples act as a "reasoning tie-breaker" between proposed solutions the model can generate. Generated solutions are not entirely irrelevant but a relatively poor interpretation of the problem. 
For example, in Figure \ref{fig:temp_2}, when stripped of its examples, the model still exhibits signs of task understanding (i.e., comparing element difference to a threshold, iterating over elements). However, combining these logic parts in a meaningful manner is complex enough that the model requires additional examples to filter out faulty strategies. We associate such effects with poor reasoning abilities.
\subsection{Framework}
\label{sec:framework}
The first step involves splitting a coding challenge into the three \textit{Blocks of Influence}. For this purpose, we utilize a regular expression module that searches for common patterns of each block's start or end. (e.g., ~\nb: "def (...):", ~\db: " or """, ~\eb: "Examples:" or $> / \gg$ followed by usage of the function name).

As the next step, the \db~ is further analyzed to identify possible hinting keywords. Ideally, we are interested in unigrams or bigrams that provide excess information towards completing the coding task. For keyword identification, we use KeyBert~\citep{grootendorst2020keybert}, an LLM tasked to perform keyword extraction and word similarity. We proceed to fine-tune KeyBert on the open-source CodeParrot dataset \cite{codeparrot} so that more code-specific suggestions are provided. 
For each candidate keyword, we calculate its embedding similarity with the set of words: {\small[Python, Programming, Code, Variable]}, again through KeyBert. Words with cosine similarity scores under 0.7 for all the items of the set are unrelated to coding and thus filtered out. 
However, carelessly removing keywords can lead to non-interesting drops in performance associated with removing crucial information rather than hinting effects. Thus, an additional context-aware filtering stage is employed to validate that any information lost can be retrieved from the remaining coding challenge. 

During this stage, we compute each candidate keyword's embedding similarity with every non-potential keyword token. The keyword is marked as valid for removal if at least one "close" word is identified. Again, we consider "close" keywords with a similarity score larger than 0.7. If a keyword exists in multiple locations, the first instance is not marked as valid for removal, while the rest are.
When a keyword happens to be an argument type (i.e., list, integer, tuple), we additionally look for instances of that type in the examples or name block. In case of a match, the keyword is safe for removal. Equivalent information already exists in the context.
As the final step, we chose between the following transformations:
\newline\textbf{Drop one.} Removes one of the provided keywords from the \db. The transformation is repeated $N$ times where $N$ is the number of identified keywords. 
\newline\textbf{Drop all.} Removes all the provided keywords simultaneously from the \db
\newline\textbf{Drop examples.} Removes all the provided examples from the \eb. 
\newline\textbf{Anonymize.} Replaces the function name with an arbitrary token. We use \textit{"func"} in our experiments. Note that the function name is also replaced in the provided examples, so no information leak occurs. We also tested whether the choice of \textit{"func"} may potentially bear some intrinsic adversarial effect associated with the training data. We experimented with other word choice replacements (\textit{"action"},\textit{"do\_stuff"}, \textit{"XYZ"}) and got the same results. 
Furthermore, we identified instances where the function name, although closely correlated to the task at hand, if it was to be taken as the sole source of information, could instead be misleading, signifying the need for proper context understanding by the tested models (See Appendix \ref{func_name}).

For example, let us use our framework on the challenge presented in Figure \ref{fig:regions}. At the first stage, KeyBert would have identified the following keywords: [Reverse, list, return, second]. Among these, the word \textit{second} does not pass the first filtering stage with over 0.7 similarity score against our set. In the second stage, each word would be compared against all the existing tokens. \textit{Reverse} and \textit{return} will not be associated with other tokens. \textit{List} will be identified in the function name and input argument type. Also, since \textit{list} is also a python keyword, it will be matched against the list type of the input given in the examples. This leaves \textit{list} as the only available keyword for removal. If keyword drop would be combined with anonymization, the drop would still be valid since the information would still be available in the examples and input type.

These transformations test the hypotheses we associate with each block, as presented in Section 
 \ref{subsec4_1}. Removing possible hints leads to performance drops between the original and modified challenges, revealing underlying biases in the models' logic
Arguably, any of our suggested transformations can destroy local semantics. However, we take significant measures to ensure that global semantics is preserved and enough information exists towards its solution.
This is also why we refrain from performing simultaneous transformations in the \eb~and \db, or all of the \textit{Blocks of Influence} together; a model stripped of all necessary information cannot generate a proper solution.
To quantify the possible degree of ambiguity our transformations introduce, we employ the LM critic test, inspired by the work of ~\cite{lm-critic, bifi}:
We collect a random sample of 200 coding challenges from the HumanEval and MBPP. Each challenge is then transformed according to the methods presented in Table \ref{lm_study}. Afterwards, for both the original and every modified version of a challenge, we calculate their log probability score using a large language model. 
The core idea is that the model will act as a soft critic, ranking model inputs by their overall plausibility. Modified inputs that seem "off" to the critic and are partially understood will be assigned a log probability score far lower than the unmodified ones. 
Since this criterion is based on local neighborhood optimality, only moderate changes are allowed between the challenges under comparison. For example, two completely different but syntactically and semantically correct text snippets can have similar log probability scores. During their comparison, however, we would have violated the locality assumption, and no conclusions could be drawn about their contents.
As our critic, we employ the Codex-v2 model \cite{chen2021evaluating}. We calculate log probability similarity as: $Sim = 100 - \frac{LogP_{Method} - LogP_{Original}}{LogP_{Original}}$.
\begin{table}[!ht]
\begin{center}
\resizebox{0.95\columnwidth}{!}{
\begin{tabular}{lcc}
\multicolumn{1}{c}{} & \multicolumn{2}{c}{Similarity (\%)}\\
\cmidrule(lr){2-3}
Method &  w/ CAF & w/o CAF\\
\midrule
Original &  100.0 ($\pm$ 0.0) & 100.0 ($\pm$ 0.0)\\
Anonymization &  98.5 ($\pm$ 1.2) & 98.5 ($\pm$ 1.2) \\
Drop One &   97.3  ($\pm$ 1.5) & 84.2  ($\pm$ 2.2)   \\
Drop All &   95.3 ($\pm$ 1.9) & 80.3 ($\pm$ 2.8)  \\
Anonymization + Drop One &   95.8 ($\pm$ 1.4) & 80.9  ($\pm$ 2.3)\\
Anonymization + Drop All &   94.6 ($\pm$ 2.3) &   78.4 ($\pm$ 3.1)\\
\bottomrule
\end{tabular}
}
\caption{\small{Similarity scores for different methods of our framework with (w/) and without (w/o) the proposed context-aware filtering mechanism (CAF). Results of 200 samples are presented.}}\label{lm_study}
\end{center}
\end{table}
\begin{table*}[!ht]
\begin{center}
\small
\caption{\small{Model results on Human Eval and MBPP.}}
\label{qual_table1}
\resizebox{2\columnwidth}{!}{
\begin{tabular}{l p{1cm} p{1.2cm} p{1cm} p{1.2cm} p{1cm} p{1.2cm} p{1cm} p{1.2cm} p{1cm} p{1.2cm} p{1cm} p{1.2cm} p{1cm}}
\toprule
\multicolumn{1}{c}{} & \multicolumn{4}{c}{Codeparrot (1.5B)} & \multicolumn{4}{c}{Incoder (1.6B) } &
\multicolumn{4}{c}{CodeGen-Mono (6B)}\\
\multicolumn{1}{c}{}  & \multicolumn{2}{c}{Human Eval} & \multicolumn{2}{c}{MBPP} & \multicolumn{2}{c}{Human Eval} & \multicolumn{2}{c}{MBPP}& \multicolumn{2}{c}{Human Eval} & \multicolumn{2}{c}{MBPP}\\
\cmidrule(lr){2-3} \cmidrule(lr){4-5} \cmidrule(lr){6-7} \cmidrule(lr){8-9} \cmidrule(lr){10-11} \cmidrule(lr){12-13}

Method\newline & Pass@1 (T=0.2) & Pass@100 (T=0.8) & Pass@1 (T=0.2) & Pass@100 (T=0.8) & Pass@1 (T=0.2) & Pass@100 (T=0.8) & Pass@1 (T=0.2) & Pass@100 (T=0.8) & Pass@1 (T=0.2) & Pass@100 (T=0.8) & Pass@1 (T=0.2) & Pass@100 (T=0.8)\\
\toprule
Original      &      4.1  & 17.8   & 6.1  &  31.2 &   11.3     &  24.2  &  14.6   &  56.7  &   26.1 	& 65.8	& 42.3	& 77.3\\  
Drop One       &     3.9  &  13.2  &  4.2 &  26.8 &    10.5  &  22.3  &  11.5 &  45.4 &   18.4 &	39.3	& 25.2	& 65.7  \\
Drop All       &     3.6  &  11.1  &  3.9 &  21.7 &     9.7  & 17.6   &  12.8 &  42.1 &  	13.9 &	34.8	& 22.4	& 57.7  \\
Drop Ex       &     3.7  & 14.3   &  5.3 &   27.5 &     11.3  &  22.2  &  14.4 &  43.8 &  20.4 & 	42.3 & 	27.2 & 	61.7 \\
Anon       &     3.8  & 12.5   &  4.7 &  23.2 &     9.1  & 21.8   &  11.3 &   45.2  &  18.2	 & 37.3	  &24.0  &	65.6  \\
Anon+Drop One       &     3.3  &  9.5  &  3.9 &  20.2 &     7.4  & 21.5  &  10.5 &   44.9  & 12.6	  &24.6	  &15.8	  &58.6   \\
Anon+Drop All        &     2.1  & 8.9  & 3.9 &   17.9 &     6.3  & 17.5  &  8.0 &   41.3 &   11.5	 & 23.1	 & 14.9	 & 46.3 \\
Anon+Drop Ex      &     3.7  & 11.8   &   4.6 &  22.8  &    8.7 & 21.3   &  11.2 &  43.5  &  16.0	&28.3	&18.2&	60.7 \\
\end{tabular}
}
\small
\resizebox{2\columnwidth}{!}{%
\begin{tabular}{l p{1cm} p{1.2cm} p{1cm} p{1.2cm} p{1cm} p{1.2cm} p{1cm} p{1.2cm} p{1cm} p{1.2cm} p{1cm} p{1.2cm} p{1cm}}
\toprule
\multicolumn{1}{c}{} & \multicolumn{4}{c}{Incoder (6B)} & \multicolumn{4}{c}{Codex (v2)} &
\multicolumn{4}{c}{Bloom (176B)}\\
\multicolumn{1}{c}{}  & \multicolumn{2}{c}{Human Eval} & \multicolumn{2}{c}{MBPP} & \multicolumn{2}{c}{Human Eval} & \multicolumn{2}{c}{MBPP}& \multicolumn{2}{c}{Human Eval} & \multicolumn{2}{c}{MBPP}\\
\cmidrule(lr){2-3} \cmidrule(lr){4-5} \cmidrule(lr){6-7} \cmidrule(lr){8-9} \cmidrule(lr){10-11} \cmidrule(lr){12-13}

Method\newline & Pass@1 (T=0.2) & Pass@100 (T=0.8) & Pass@1 (T=0.2) & Pass@100 (T=0.8) & Pass@1 (T=0.2) & Pass@100 (T=0.8) & Pass@1 (T=0.2) & Pass@100 (T=0.8) & Pass@1 (T=0.2) & Pass@100 (T=0.8) & Pass@1 (T=0.2) & Pass@100 (T=0.8)\\
\toprule
Original       &      15.2  & 47.0  &  19.4  &  65.1 &      49.4  &   91.4 &  60.1  &  86.3  &      16.4  &   57.2 &  20.8  & 62.4\\  
Drop One       &    12.1  & 35.3  &  18.9 &   52.6 &    36.0   &  86.2 & 56.0 & 79.2 &    12.8   &  48.6 & 15.8 & 51.4\\
Drop All       &     10.2  & 28.2   &  15.6 &  47.0  &  37.1     & 73.7  & 52.1 & 69.5  &  11.5     &  40.2  & 14.2 & 44.4 \\
Drop Ex       &     12.7  &  29.5  &  17.4 &   50.3 &  41.4     & 81.0   & 48.8 & 70.7  &  15.2     & 43.3  & 15.8 & 50.1\\
Anon       &     11.6  &  32.9  &  14.8 &  50.7 &     44.5  &  90.4     &  57.9 &  81.7  &     14.0  &  48.3   &  15.1 &  51.2\\
Anon+Drop One       &     8.1 & 30.6   &  13.5 &  46.7 &     29.8  & 74.4  &  51.2   &  69.5  &   12.8  & 41.9 &  13.6    &  46.8\\
Anon+Drop All        &     7.5 & 25.2  &  11.2 &  38.9  &   24.2  & 68.7  &   47.2    &  63.8 &  10.3  & 36.8  & 12.6  & 38.4\\
Anon+Drop Ex      &     11.2  & 28.1   &   14.5 &  50.2 &   34.1    & 72.5 &  42.6     & 70.5 &   14.0    & 39.8  &  14.3    & 47.8\\
\bottomrule

\end{tabular}
}
\end{center}
\end{table*}
\newline Table \ref{lm_study} shows that our transformations do not introduce drastic changes to the coding challenge. Even in the most aggressive transformation of \textit{Anonymization + Drop All}, the critic assigns over $94\%$ similarity between code challenges affected by it versus their original form. For comparison, removing the context-aware filtering stage, leads to only $78\%$ similarity in the case of \textit{Anonymization + Drop All} transformation.
We believe this is a fair indicator that the tested models observe inputs of similar quality and comprehensibility during our experiments.
Note that we omit results for the \textit{Drop Examples} method. In this case, the log probabilities will significantly change since we remove many tokens, which violates the method's locality prerequisite. 
\section{Experiments}
\label{sec:exp}
\subsection{Results on Block Transformations}
The main results of our experiments are presented in Table \ref{qual_table1}.  Despite their simplicity, our transformations cause consistent drops in performance across different model sizes on both datasets.\footnote{We present a full table of results, including Codeparrot (110M), CodeGen(350M), Bloom(1.7B) and Codex(v1) in Appendix \ref{sec:quantapx}}
Mere anonymization causes drops of 19\% on average in both Pass@1 and Pass@100 metrics, validating our claims of memorization effects. Single (\droponeI) and full keyword removal (\dropallI) reduce models' performance by 15\% and 22\% on average, suggesting their inability to deduct the missing context from \nb ~and \eb. Instead, models rely on generating arbitrary, commonly used snippets that vaguely fit for the task.
Especially interesting are the cases of \dropexI ~and \anonplusdropexI, with 15\% and 25\% average drops. Both transformations remove the information provided by the docstring examples, with the latter having the additional restriction of an anonymized function. With the \db ~unmodified in both cases, these transformations target the models' abilities to create solutions based on their natural language understanding. 
The combination of anonymization with the drop of all keywords (\anonplusdropallI) seems to be the most challenging transformation overall, with drops of approximately 40\%. Its primary purpose is to assess the model's capability of deducting the missing context of the \db ~by only observing patterns in the examples. 
\begin{table*}
\begin{center}
\caption{\small {HumanEval results of fine-tuning Codeparrot on the MBPP dataset with (A) or with no (NA) augmentations: Regular finetuning does not contribute to bias removal, achieving similar results against the perturbations. However, our suggested augmentations lead to higher model performance, especially in the pass@100 metric. The average of 15 runs is presented. Bold marks statistically significant improvements under the T-Test (Before versus After-A) with $a=0.95$.}}
\label{mini_table_min}
\resizebox{1.98\columnwidth}{!}{
\begin{tabular}{c c c c c c c c c c c c c c}
\toprule
\multicolumn{1}{c}{} & \multicolumn{4}{c}{Codeparrot - 110M} & \multicolumn{4}{c}{Codeparrot - 350M} & \multicolumn{4}{c}{Codeparrot - 1.5B} \\
\multicolumn{1}{c}{} & \multicolumn{2}{c}{Pass@1 (T=0.2)} & \multicolumn{2}{c}{Pass@100 (T=0.8)} & \multicolumn{2}{c}{Pass@1 (T=0.2)} & \multicolumn{2}{c}{Pass@100 (T=0.8)} & \multicolumn{2}{c}{Pass@1 (T=0.2)} & \multicolumn{2}{c}{Pass@100 (T=0.8)} \\
\cmidrule(lr){2-3} \cmidrule(lr){4-5} \cmidrule(lr){6-7} \cmidrule(lr){8-9} \cmidrule(lr){10-11} \cmidrule(lr){12-13}
\multicolumn{1}{c}{} & \multicolumn{1}{c}{Before} & \multicolumn{1}{c}{After} & \multicolumn{1}{c}{Before} & \multicolumn{1}{c}{After} &  \multicolumn{1}{c}{Before} & \multicolumn{1}{c}{After} &  \multicolumn{1}{c}{Before} & \multicolumn{1}{c}{After} &  \multicolumn{1}{c}{Before} & \multicolumn{1}{c}{After} & \multicolumn{1}{c}{Before} & \multicolumn{1}{c}{After}\\

\multicolumn{1}{c}{Method} &
\multicolumn{1}{c}{} & \multicolumn{1}{c}{NA / A} &
\multicolumn{1}{c}{} & \multicolumn{1}{c}{NA / A} &
\multicolumn{1}{c}{} & \multicolumn{1}{c}{NA / A} &
\multicolumn{1}{c}{} & \multicolumn{1}{c}{NA / A} &
\multicolumn{1}{c}{} & \multicolumn{1}{c}{NA / A} &
\multicolumn{1}{c}{} & \multicolumn{1}{c}{NA / A}\\

\toprule
Original
& 3.8 &         3.7 / 3.7
& 12.7 &         12.1     / 12.1 
& 3.8 &           3.7    / 3.7 
& 13.9 &         13.7    / 13.7 
& 4.1 &          4.1     / 4.1
& 17.8 &         17.8    / 17.8\\

Drop One
& 3.3 &          3.2     / \textbf{3.6}
& 9.7 &           9.7   / \textbf{10.4}  &
3.3 &           3.3    / \textbf{3.6} &
11.9 &          11.9     /  \textbf{12.3} &
3.9 &            3.9    / \textbf{4.0} &
13.2 &           13.2   / \textbf{14.1} \\

Drop All
& 3.1 &          3.1     / 3.1 &
7.2 &            7.2     /  \textbf{7.9} &
3.2 &             3.2    / 3.2 &
10.1 &            10.0    /  \textbf{10.7} &
3.6 &             3.6     /  3.7 &
11.1 &           11.1    / \textbf{12.3}\\

Drop Ex
& 3.8 &          3.7     /  3.7 &
9.9 &             9.9   /  \textbf{10.2} &
3.8 &             3.8   /   3.7 &
12.9 &           12.9   /  12.9 &
3.7 &          3.7    /  3.7 & 
14.3 &          14.3  /  \textbf{15.1}\\

Anon
& 3.4 &           3.4      /  \textbf{3.5} &
8.7 &             8.7   / \textbf{9.1} & 
3.6  &            3.6    /   3.6 &
11.6 &             11.6  / \textbf{12.2} &
3.8  &             3.8   /   3.9 &
12.5 &            12.5   /  \textbf{13.8}\\

Anon+Drop One
& 3.0 &          2.8  /   \textbf{3.4} &
7.5 &          7.5    /   \textbf{7.9} &
3.0 &           2.8    /   \textbf{3.5} &
8.2 &           8.2    /  \textbf{9.4} &
3.3 &          3.3     /  \textbf{3.5} & 
9.5 &           9.5     /  \textbf{10.5}\\
    
Anon+Drop All
& 1.9 &         1.9 /  2.0 &
6.9 &           6.9  /   6.9 &
2.0 &         2.0   /   \textbf{2.2} &
8.1 &          8.0  /    8.3 &
2.1 &          2.1    /  \textbf{2.4} &
8.9 &          8.8  /   9.4\\
    
Anon+Drop Ex
& 3.4 &        3.3    /   3.4 & 
8.7 &          8.7  /  \textbf{9.0} &
3.6 &         3.6  /    3.6 & 
10.7 &         10.7  /   \textbf{11.8} &
3.7 &         3.7   /   3.7 &
11.8 &         11.8  /  \textbf{13.7}\\
\bottomrule
\end{tabular}
}
\end{center}
\end{table*}
These observations suggest a clear model preference over its sources of information, with the task description being the primary one. Thus, when a model exhausts its ability to understand the task, it exploits similarities of the function name with previously seen code solutions. Simultaneously, the model's reasoning relies on the example demonstrations, which, as seen from (\anonplusdropallI), are not always able to provide clear directives.
\subsection{Towards Bias Mitigation}
\label{sec:defense}
Inspired by the field of adversarial training, we decided to investigate the effects of using our framework transformations as training augmentations. To this end, we apply our framework to examples of the MBPP challenge and use them as a fine-tuning dataset for three different Codeparrot 
models. We use HumanEval as our test dataset, which bears no overlap with the MBPP. In this way, our models have not seen examples of the test set during their training or fine-tuning steps. In Table \ref{mini_table_min},
we compare the results of our models before and after fine-tuning. Models benefit from the introduction of augmented examples and partially recover from failure modes caused by the need to rely on hints. The larger the model, the more its abilities benefit. We believe this effect is closely related to large language models' scaling reasoning capabilities and their parameter size. The need to rely on hints can be attributed to low data quality or lack of task-specific inductive biases. However, the capacity to properly understand coding tasks is undoubtedly there. To improve the code generation abilities of models, we thus suggest exposing them to challenges that push their deductive and reasoning abilities.
We decided to repeat the experiments, but without including any of our data augmentation techniques during fine-tuning. We observe that under this setup, models do not exhibit any significant improvement against our method's perturbations. Our suggested data augmentations that push the reasoning limits of the models are thus a valid alternative to simple fine-tuning.
\subsection{Effects of Longer Context}\label{sec:longcont}
When causally training on coding datasets, models condition on multiple functions and declarations in the same file. The input is a conglomerate of rapidly changing contexts, with each function or class being a self-contained entity.
Subsequently, a model is accustomed to localizing its focus when trained on such data. As an extension to our previous experiment, we measure the effects of using a long description dataset, DMCC, as a fine-tuning target. By training on long descriptions of natural language, we promote the context-deducting skills of the model under test.
\begin{table}[!hb]
\begin{center}
\caption{\small{HumanEval results of fine-tuning Bloom (560M) on the modified MBPP and long-description DMCC dataset with (A) or without (NA) augmentations: Model exhibits increased performance under the combined augmentation schema against perturbations that challenge language understanding. The average of 15 runs is presented. Bold marks statistically significant improvements under the T-Test (Before versus $+$DMCC-A) with $a=0.95$.}}
\label{long_context}
\resizebox{1\columnwidth}{!}{
\begin{tabular}{c c c c c c c c}
\toprule
\multicolumn{1}{c}{} & \multicolumn{3}{c}{Pass@1 (T=0.2)} & \multicolumn{3}{c}{Pass@100 (T=0.8)} \\
\cmidrule(lr){2-4} \cmidrule(lr){5-7}
\multicolumn{1}{c}{} & \multicolumn{1}{c}{Before} & \multicolumn{1}{c}{+MBPP}  & \multicolumn{1}{c}{+DMCC} & \multicolumn{1}{c}{Before} & \multicolumn{1}{c}{+MBPP} &  \multicolumn{1}{c}{+DMCC}\\

\multicolumn{1}{c}{Method} & \multicolumn{1}{c}{} &
\multicolumn{1}{c}{NA / A} &
\multicolumn{1}{c}{NA / A} & \multicolumn{1}{c}{} &
\multicolumn{1}{c}{NA / A} & \multicolumn{1}{c}{NA / A}\\
\toprule
Original          
&  3.7 &  3.6 / 3.6 &  3.6 / 3.6 
& 12.1  &   12.1 / 12.1 &  12.0 / 12.0\\
Drop One           
&  3.1 &   3.1 / 3.6  &   3.1  / \textbf{3.6} 
& 10.3 &   10.3 /  10.9 &  10.3 / \textbf{10.9}\\
Drop All           
&  2.4 &   2.3  / 2.4 &   2.3 / \textbf{2.9} 
& 9.2 &   9.1  / 9.1 &    9.1 / \textbf{9.7}\\
Drop Ex            
&   3.0 &  3.0 / 3.0  &   3.0   / 3.0  
& 11.0 &   11.0 / 11.3 &   11.0 / \textbf{11.5}\\
Anon               
&  2.5 &   2.5 / 3.0 &    2.6 / \textbf{3.6} 
& 10.7  &  10.7 / 10.9 &    10.8 / \textbf{11.3}\\
Anon+Drop One     
&   1.9 &  1.9 / 2.3 &   1.9 / \textbf{2.4}  
& 7.8 &    7.8 / 9.1 &    7.8 / \textbf{9.7}\\
Anon+Drop All     
&   1.8 &   1.8 /  1.8 &    1.8 / \textbf{2.3} 
& 7.0 &    7.0 / 7.2 &   7.0 / \textbf{8.3}\\
Anon+Drop Ex      
&   2.4 &   2.4 / 2.9 &  2.4  / \textbf{3.0}  
& 9.7  &    9.7 / 10.3 &   9.7 / \textbf{11.4}\\
\bottomrule
\end{tabular}
}
\end{center}
\end{table}
A model able to widen its focus can avoid distractions caused by missing keywords. Efficient context understanding will replace  not rely heavily on internal biases. We choose Bloom as the model under test since it was not explicitly tuned for code generation but rather general language understanding. 
In Table \ref{long_context}, we present results of fine-tuning on MBPP, modified by our framework. We observe similar performance improvements as in Table \ref{mini_table_min}. We experiment again, this time combining both MBPP and  DMCC examples. We show that incorporating examples of more extended context leads to even better performance against transformations targeting the \db ~and language understanding. Similar experiments were conducted with the CodeParrot variants but were unfruitful. We attribute this to the restricted focus regarding training data (exclusively Python3 code) and architectural differences between the models. We believe that the merging benefits of our two proposed setups can serve as an interesting direction towards model resilience in code generation scenarios.

\section{Conclusions}
We present a simple approach to isolate cues and benchmark the reasoning  of code generation models through input-level transformations.
\begin{figure}[!ht]
% \minipage{\columnwidth}
% \begin{minted}
% [frame=single,
% framesep=1mm, 
% fontsize=\tiny,
% breaklines,
% highlightlines={7}
% ]
% {python}
% from typing import List
% def string_xor(a:str, b:str) -> str:
%     "Input is two strings a and b consisting only of 1s and 0s. Perform binary XOR on these inputs and return the result as a string.
%     Examples:
%     >>> string_xor('010','110')
%     '100'
%     return ''.join([str(int(a,2) ^ int(b,2)) for a,b in zip(a,b)])
% \end{minted}
% \endminipage \hfill
% \minipage{\columnwidth}
% \begin{minted}
% [frame=single,
% framesep=1mm, 
% fontsize=\tiny,
% breaklines,
% highlightlines={7},
% highlightcolor={pink}
% ]
% {python}
% from typing import List
% def string_xor(a:str, b:str) -> str:
%     "Input is a and b consisting only of 1s and 0s. Perform  XOR on these inputs and return the result.
%     Examples:
%     >>> string_xor('010','110')
%     '100'
%     return ''.join([str(int(a) ^ int(b))])
% \end{minted}
% \endminipage
\includegraphics[width=\linewidth]{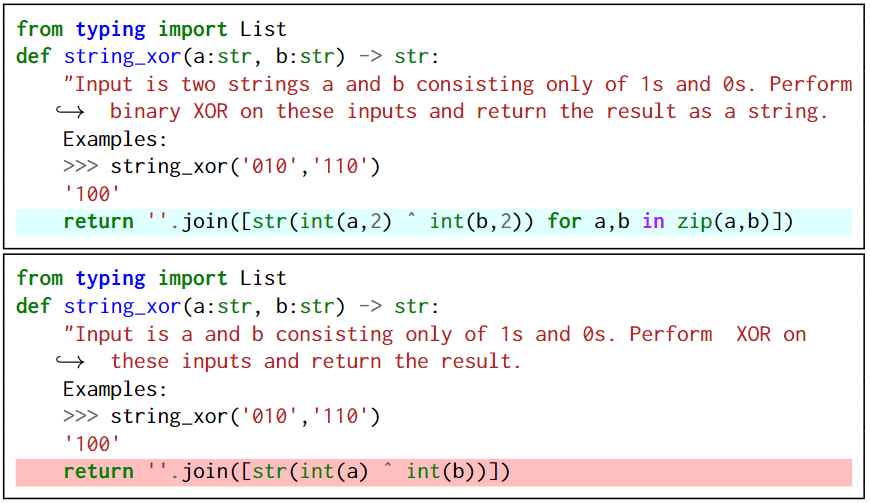}
\caption{\footnotesize{Keyword hinting (\dropallI ~ / Bloom 175B): After the removal of keywords, the context remains intact: The \textit{two strings} keyword can be assumed by observing the function arguments, and the \textit{binary/string} keywords by the examples and return type signature of the function. Nevertheless, the model fails to generate a correct solution (bottom).}}
\label{fig:temp_3}
\end{figure}
Our method treats code examples as a combination of three blocks, each providing different cues to the model. We show that minor transformations can lead models to failure, signifying the existence of biases. Our framework can automatically identify and remove keywords responsible for indirect hinting.
We show that popular models with solid results on challenging coding challenges are susceptible to our tests, with their performance degrading noticeably. 
Moreover, we studied the effects of utilizing our proposed transformations during the fine-tuning of a model.
Models can benefit from our proposed changes, with the effect proportional to their parameter size. We believe that, despite their success, code generation systems with LLMs as backbones inherit some of their biases and modes of failure. Training on structured and well-documented code, combined with our proposed techniques, is a promising direction towards reliable code generation. Although an ideal fit for competition-style challenges, our method can be extended to support less formatted high-quality codebases (e.g. GitHub repositories).
For a short analysis see Section \ref{sec:opensource} of the Appendix.
\begin{figure}[!ht]
% \minipage{\columnwidth}
% \begin{minted}
% [frame=single,
% framesep=1mm, 
% fontsize=\tiny,
% breaklines,
% highlightlines={10,11,12,13,14}
% ]
% {python}
% def pairs_sum_to_zero(l):
%     "Pairs_sum_to_zero takes a list of integers as input. It returns True if there are two distinct elements in the list that sum to zero, and False otherwise."
%     Examples:
%     >>> pairs_sum_to_zero([1,3,5,8])
%     False
%     >>> pairs_sum_to_zero([1,3,-2,1])
%     False
%     >>> pairs_sum_to_zero([2,4,-5,3,5,7])
%     True
%     for i in range(len(l)):
%         for j in range(i+1, len(l)):
%             if l[i] + l[j] == 0:
%                 return True
%     return False
% \end{minted}
% \endminipage \hfill
% \minipage{\columnwidth}
% \begin{minted}
% [frame=single,
% framesep=1mm, 
% fontsize=\tiny,
% breaklines,
% highlightlines={3},
% highlightcolor={pink}
% ]
% {python}
% def func(l):
%     "Func takes a list of integers as input. It returns True if there are two distinct elements in the list that sum to zero, and False otherwise."
%     return any(sum(x) == 0 for x in l)
% \end{minted}
% \endminipage
\includegraphics[width=\linewidth]{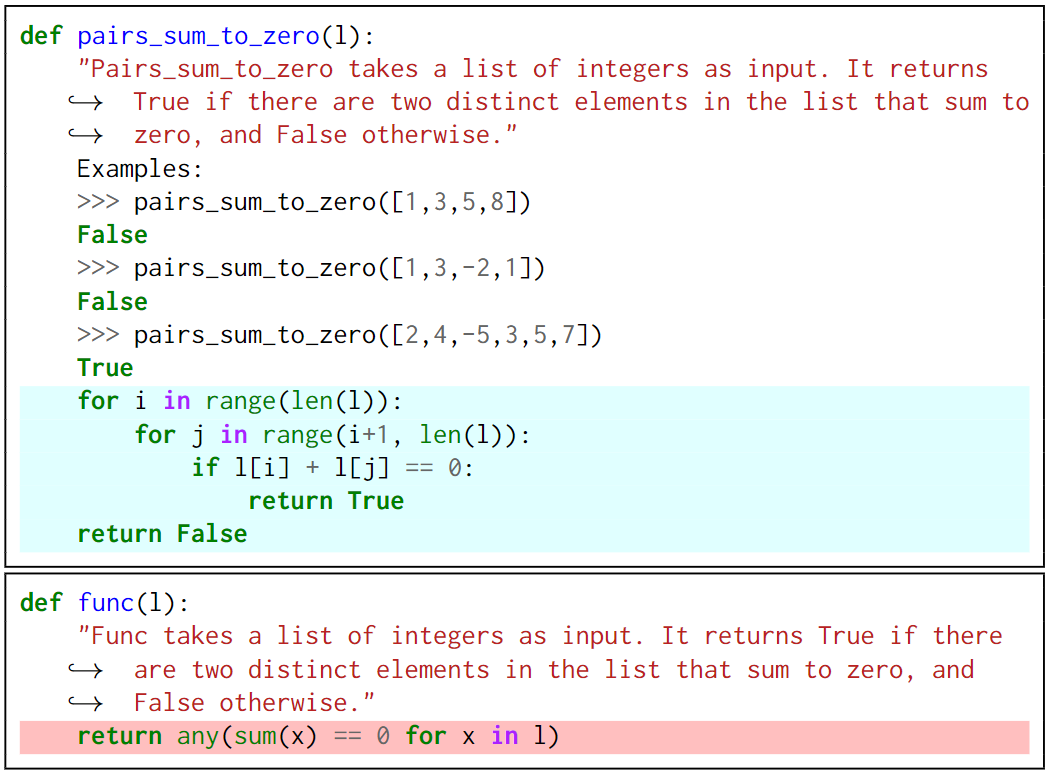}
\caption{\footnotesize{\anonplusdropexI ~ / Incoder 6B: Using only the problem description, the model creates partially informed subparts (any derives from \textit{"if there are"}, sum(x) == 0 from \textit{"sum to zero"}, and for x in l from \textit{"elements in the list"}) that are not combined correctly to solve the task (bottom), signifying that hints from the function name/examples were used in the correct solution (top).}}
\label{fig:temp_4}
\end{figure}

\section{Limitations}
Some limitations and possible research directions exist in our work. Our study focuses on the Python3 programming language, with many coding challenges existing in different popular choices (e.g., C, C++, Java, Scala). Although the Blocks of Influence identification mechanism could be easily adapted to each case, an off-the-shelf application of our framework in another language would lead to errors. 

Similarly, the framework assumes that each coding challenge will be in a "competition-style" format, meaning that a proper problem description, in-docstring examples, and each input types are present for each example. In Appendix Section \ref{sec:opensource}, we present how an adaptation to less formatted codebases would be possible, but for now, we leave it as a future investigation.

Finally, there is no guarantee that the improved performance against the suggested perturbations reflects an equivalent performance increase in real-world code assistant applications. Real-time coding suggestions and completions that are more user aligned are out of the scope of this work.

\section{Risks and Ethical Considerations}
Our research aims to discover and remove biases in code-generation scenarios through adversarial intervention. However, we acknowledge that insecure or malicious code can still be generated after finetuning with our suggested augmentations.
Furthermore, our work is focused only on cognitive biases that affect the reasoning and logic behind the coding process of large language models. Social biases and stereotypes can still appear when general-purpose LLMs such as Codex or Bloom are used in typical text generation scenarios. Signs of robustness against our methods are not to be confused with indicators of other forms of biases not existent.
\section*{Acknowledgements}
All experiments were performed using the Entropy cluster funded by
NVIDIA, Intel, the Polish National Science Center grant
UMO-2017/26/E/ST6/00622 and ERC Starting Grant TOTAL. The work of
Spyridon Mouselinos and Henryk Michalewski was supported by the Polish
National Science Center grant UMO-2018/29/B/ST6/02959.

\bibliography{acl}
\bibliographystyle{acl_natbib}
\clearpage
\onecolumn

\section{Appendix}
\subsection{Extension to open-source code}
\label{sec:opensource}
Although an ideal fit for competition-style challenges, our method can be extended to support less formatted high-quality codebases (e.g. GitHub repositories). Large files can be broken down into individual functions/classes, each further analyzed into Blocks of Influence. In such codebases, function names should be closely relevant to their purpose. The existence of meaningful docstrings is crucial, the absence of which promotes more memorization and biases as we exhibited. Moreover, the input/output checks contained in function unit tests can be repurposed as function examples. Keywords can be chosen similarly, with the context being co-informed by both local and larger scopes.

\subsection{Information on Models and Datasets}
\label{sec:modelsanddata}
\begin{table}[!ht]
\begin{center}
\small
\resizebox{\columnwidth}{!}{
\begin{tabular}{lcccl}
Model Name &  Link  & LICENSE \\
\midrule
KeyBert \cite{grootendorst2020keybert} & https://github.com/MaartenGr/KeyBERT & MIT\\
Codeparrot \cite{tunstall2022natural}  & https://huggingface.co/codeparrot/codeparrot & Apache License 2.0    \\
InCoder \cite{fried2022incoder}  &  https://github.com/dpfried/incoder & CC-BY-NC 4.0\\
CodeGen \cite{CodeGen}  & https://github.com/salesforce/CodeGen & BSD 3-Clause\\
Bloom \cite{bloom2022} &  https://huggingface.co/bigscience/bloom & BigScience RAIL License v1.0\\
Codex-V2 \cite{chen2021evaluating}  &   https://beta.openai.com/ &  N/A\\
\bottomrule
\end{tabular}
}
\end{center}
\caption{URL and Licenses of used Models.}
\label{table:model_links}
\end{table}

\begin{table}[!ht]
\begin{center}
\small
\resizebox{\columnwidth}{!}{
\begin{tabular}{lcc}
Dataset Name &  Link & LICENSE  \\
\midrule
CodeParrot Dataset \cite{codeparrot} & https://huggingface.co/datasets/codeparrot/codeparrot-clean & Apache License 2.0\\
HumanEval \cite{chen2021evaluating}& https://github.com/openai/human-eval & MIT \\
MBPP \cite{mbpp}  &  https://github.com/google-research/google-research/tree/master/mbpp & CC BY 4.0\\
DMCC \cite{alphacode}  &  https://github.com/deepmind/code\_contests & Apache License 2.0\\
\bottomrule
\end{tabular}
}
\end{center}
\caption{URL and Licenses of used Datasets.}
\label{table:data_links}
\end{table}

\begin{table}[!htbp]
\begin{center}
\small
\resizebox{\columnwidth}{!}{
\begin{tabular}{lccccc}
Name &  $\#$Problems & $\#$Tests per Problem & Avg. desc. length & Avg. keywords \\
\midrule
HumanEval \cite{chen2021evaluating}&  164 & 8 & 449 & 4\\
MBPP \cite{mbpp}  &  1000   & 3  & 235 & 4 \\
DMCC (Train / Python3) \cite{alphacode}  &  8139   & 85  &  1480 &  9\\
\bottomrule
\end{tabular}
}
\caption{Datasets used in experiments. We present the number of problems, number of tests per problem, average length of the challenge description and average distinct keywords identified by our framework.\label{dataset_table}}
\end{center}
\end{table}
For all of our perturbation experiments, we utilize the abovementioned models, and we comply with their respective licenses and intended use (generating code completions in python3). This also stands true for Codeparrot and Bloom, for which we create fine-tuned versions. Furthermore, we do not plan to repack or redistribute any of the used datasets. We plan to release the codebase of this work as an open-source project.

\subsection{Information on Experimental Setup}
Our experimental setup consisted of 4x   NVIDIA V100 GPUs. Regarding the results of Table \ref{qual_table1}, the computing time of each table entry was influenced by: the model size, the k value of pass@k metric (number of generations), the perturbation method, and the dataset tested. Specifically for the drop one / anonymize + drop one methods, the experiment was repeated N times, where N corresponds to the number of keywords identified. This results in approximately four times slower experiments for those perturbations since in both HumanEval and MBPP, four keywords on average per problem were identified (see Table \ref{dataset_table}). API calls to Codex and Bloom models were subject to throttling limits, and waiting loops were introduced to avoid interruptions of service.
The total experiment time resulted in approximately 500 hours.

Regarding the finetuning experiments of Table \ref{mini_table_min}, we trained Codeparrot Models with the AdamW optimizer at a learning rate of 1e-5, batch size of 64, weight decay of 0.01, and constant learning rate schedule. The same hyperparameters were chosen as well in the case of the MBPP-only experiment of the Bloom Model in Table \ref{long_context}. When both MBPP and DMCC datasets were combined, a learning rate of 3e-5 and a batch size of 256 were used. The hyperparameters were chosen after a grid search on the following choices: Weight decay (0.01 / 0.0), Learning Rate: (1e-6,1e-5,3e-5,5e-5,1e-4), Schedule: (Constant, Cosine). The batch size was chosen proportionally to the overall dataset length. All models were trained with the Accelerate library \cite{accelerate} and Zero-3 \cite{zero} partitioning schema.
Regarding the training objective, we used a custom causal language modeling loss. The loss was calculated only on the generated tokens corresponding to the problem solution and not on any tokens belonging to the problem description or examples. We used a random validation split of 10\% and validation loss for all experiments as our metric for early stopping.

\subsection{Qualitative Examples}
We present examples of code generation failures caused by our framework across different models and scenarios. On each pair, the left image represents the original, unmodified challenge alongside the correctly generated solution. The right image contains the modified version of the challenge and the incorrect completion.

\begin{figure}[!ht]
% \minipage{0.49\columnwidth}
% \begin{minted}
% [frame=single,
% framesep=1mm, 
% fontsize=\tiny,
% breaklines,
% highlightlines={8,9,10,11,12,13,14,15,16,17,18,19}
% ]
% {python}
% def is_equal_to_sum_even(n):
%     "Evaluate if n can be written as the sum of exactly 4 positive even numbers."
%     Examples:
%     >>> is_equal_to_sum_even(4)
%     False
%     >>> is_equal_to_sum_even(8)
%     True
%     if n % 2 != 0:
%         return False
%     for i in range(1, n // 2 + 1):
%         for j in range(1, n // 2 + 1):
%             for k in range(1, n // 2 + 1):
%                 for l in range(1, n // 2 + 1):
%                     if i + j + k + l == n:
%                         return True
%     return False
% \end{minted}
% \endminipage\hfill
% \minipage{0.49\columnwidth}
% \begin{minted}
% [frame=single,
% framesep=1mm, 
% fontsize=\tiny,
% breaklines,
% highlightlines={8,9,10,11,12,13,14,15,16,17,18},
% highlightcolor={pink}
% ]
% {python}
% def func(n):
%     "Evaluate if n can be written as the sum of exactly 4 positive even numbers."
%     Examples:
%     >>> func(4)
%     False
%     >>> func(8)
%     True
%     if n % 2 != 0:
%         return False
%     if n < 8:
%         reuturn False
%     if n == 8:
%         return True
%     if n > 8:
%         return func(n - 2)
% \end{minted}
% \endminipage
\includegraphics[width=\linewidth]{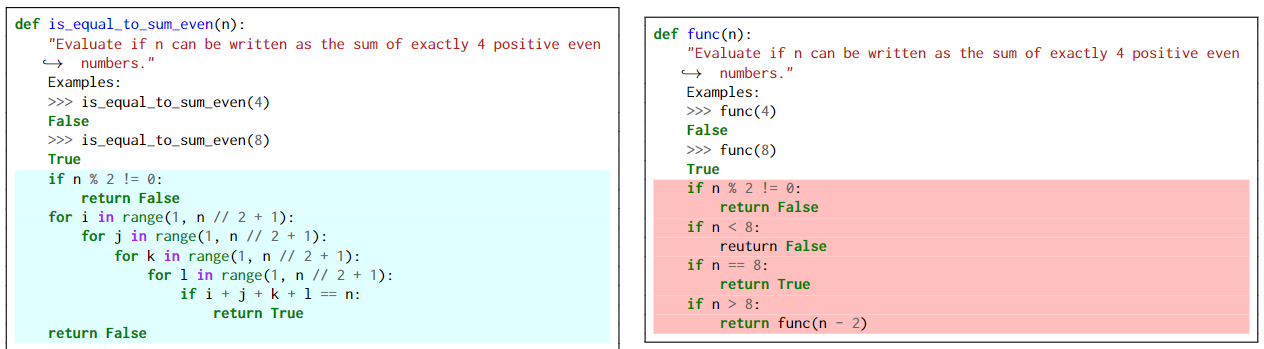}
\caption{Instance of anonymization on Codex-V1}
\label{fig:temp_1}
\end{figure}

\begin{figure*}[!ht]
% \minipage{0.49\columnwidth}
% \begin{minted}
% [frame=single,
% framesep=1mm, 
% fontsize=\scriptsize,
% breaklines,
% escapeinside=||,
% highlightlines={11,12},
% ]
% {python}
% def remove_vowels(text):
%     "Remove_vowels is a function that takes a string and returns the string without vowels"
%     >>> remove_vowels('')
%     ''
%     >>> remove_vowels('abcdef')
%     'bcdf'
%     >>> remove_vowels('aaaaa')
%     ''
%     >>> remove_vowels('aaBAAA')
%     'B'
%     vowels = 'aeiouAEIOU'
%     return ''.join([c for c in text if c not in vowels])
% \end{minted}
% \endminipage \hfill
% \minipage{0.49\columnwidth}
% \begin{minted}
% [frame=single,
% framesep=1mm, 
% fontsize=\scriptsize,
% breaklines,
% escapeinside=||,
% highlightlines={4,5,6},
% highlightcolor={pink}
% ]
% {python}
% def remove_vowels(text):
%     "Remove_vowels is a function that takes a string and returns the string without vowels"
%     vowels = ['a','e','i','o','u']
%     for vowel in vowels:
%         text = text.replace(vowel, '')
%     return text
% \end{minted}
% \endminipage
\includegraphics[width=\linewidth]{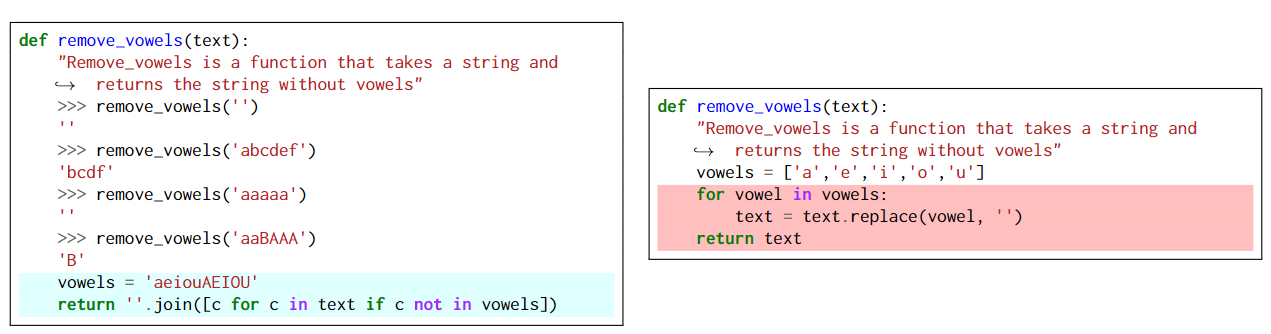}
\caption{Instance of dropping the prompt examples on Codex-V2}
\label{fig:attacks_1}
\end{figure*}

\begin{figure*}[!ht]
\includegraphics[width=\linewidth]{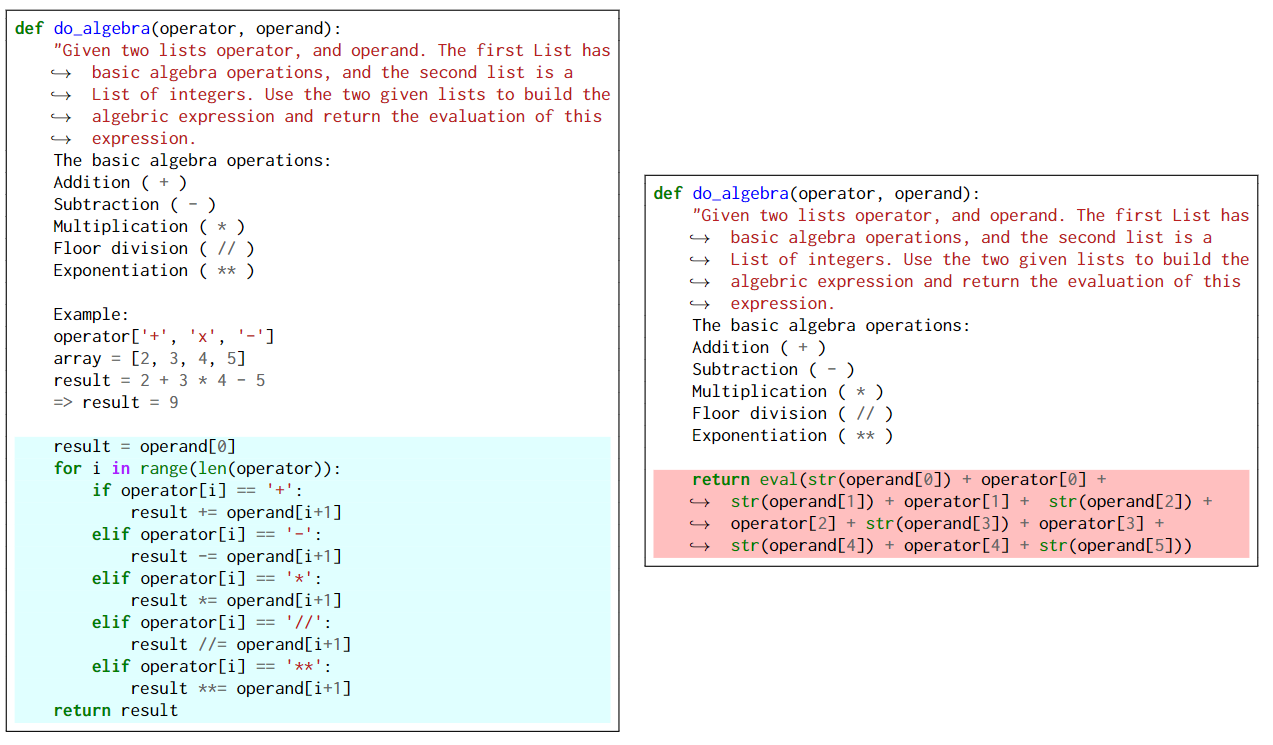}
\caption{Instance of dropping the prompt examples on CodeParrot-1.7B}
\label{fig:attacks_2}
\end{figure*}

\begin{figure*}[!ht]
% \minipage{0.49\columnwidth}
% \begin{minted}
% [frame=single,
% framesep=1mm, 
% fontsize=\scriptsize,
% breaklines,
% escapeinside=||,
% highlightlines={8},
% ]
% {python}
% def square_nums(nums):
%     "Find squares of individual elements in a list using the lambda function."
%     Examples:
%     >>> square_nums([1, 2, 3, 4, 5, 6, 7, 8, 9, 16])
%     [1, 4, 9, 16, 25, 36, 49, 64, 81, 100]
%     >>> square_nums([10,20,30])
%     [100,400,900]
%     return list(map(lambda x: x**2, nums))
% \end{minted}
% \endminipage \hfill
% \minipage{0.49\columnwidth}
% \begin{minted}
% [frame=single,
% framesep=1mm, 
% fontsize=\scriptsize,
% breaklines,
% escapeinside=||,
% highlightlines={8,9,10,11,12,13,14,15,16,17},
% highlightcolor={pink}
% ]
% {python}
% def func(nums):
%     "Find squares of individual elements using the lambda function."
%     Examples:
%     >>> func([1, 2, 3, 4, 5, 6, 7, 8, 9, 16])
%     [1, 4, 9, 16, 25, 36, 49, 64, 81, 100]
%     >>> func([10,20,30])
%     [100,400,900]
%     s=0
%     for i in range(len(nums)):
%         if nums[i] == 0:
%             s+=1
%         elif s % 2 == 0:
%             s+=1
%         else:
%             s+=1
%     return s
% \end{minted}
% \endminipage
\includegraphics[width=\linewidth]{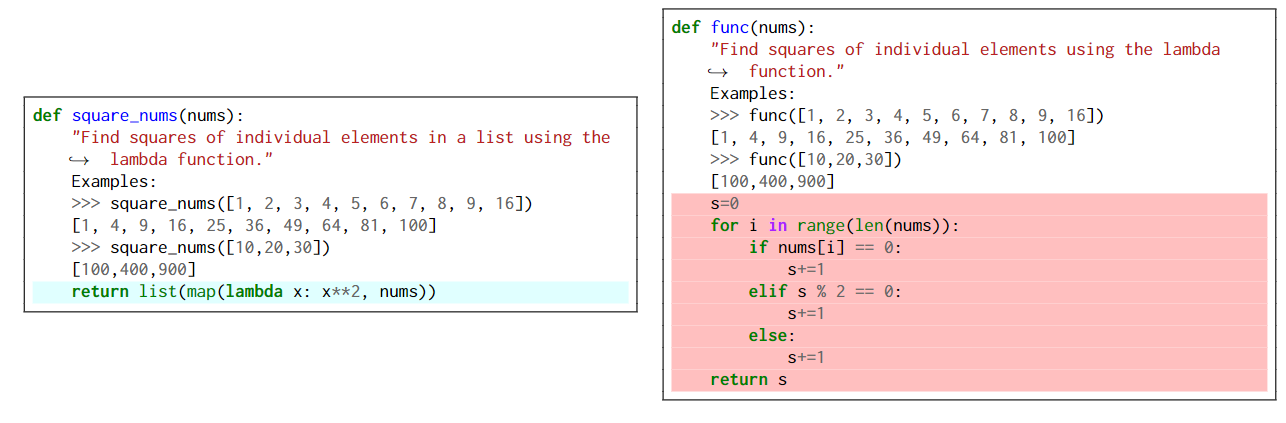}
\caption{Instance of anonymization and keyword drop on Bloom-1.7B}
\label{fig:attacks_3}
\end{figure*}

\begin{figure*}[!ht]
% \minipage{0.49\columnwidth}
% \begin{minted}
% [frame=single,
% framesep=1mm, 
% fontsize=\scriptsize,
% breaklines,
% escapeinside=||,
% highlightlines={11},
% ]
% {python}
% def iscube(a):
%     "Write a function that takes an integer a and returns True if this integer is a cube of some integer number."
%     Examples:
%     iscube(1)   ==> True
%     iscube(2)   ==> False
%     iscube(-1)  ==> True
%     iscube(64)  ==> True
%     iscube(0)   ==> True
%     iscube(188) ==> False
    
%     return a == 0 or a == 1 or (a > 0 and a**(1/3) == int(a**(1/3)))
% \end{minted}
% \endminipage \hfill
% \minipage{0.49\columnwidth}
% \begin{minted}
% [frame=single,
% framesep=1mm, 
% fontsize=\scriptsize,
% breaklines,
% escapeinside=||,
% highlightlines={,11,12,13,14,15,16,17},
% highlightcolor={pink}
% ]
% {python}
% def iscube(a):
%     "Write a function that takes a and returns True if a cube of some integer number."
%     Examples:
%     iscube(1)   ==> True
%     iscube(2)   ==> False
%     iscube(-1)  ==> True
%     iscube(64)  ==> True
%     iscube(0)   ==> True
%     iscube(188) ==> False
    
%     if (a ** 3 == a):
%         return True
%     else:
%         return False
% \end{minted}
% \endminipage
\includegraphics[width=\linewidth]{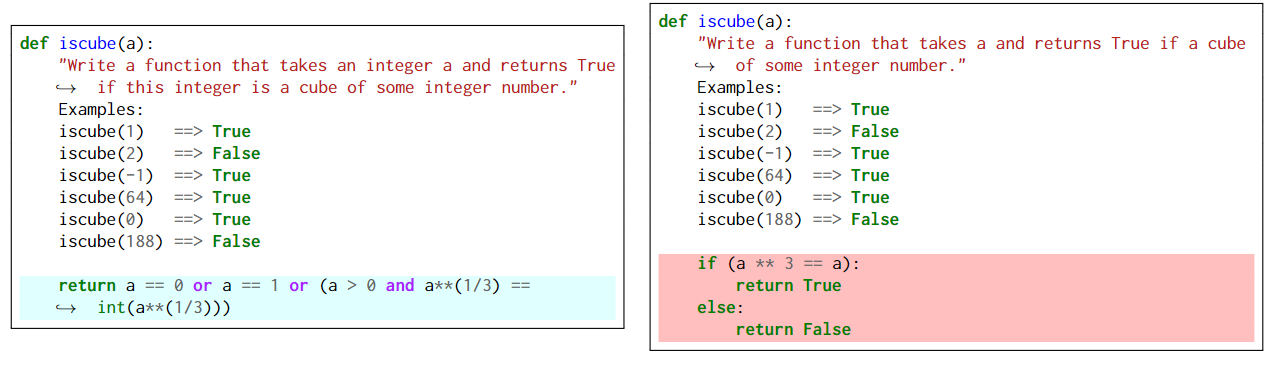}
\caption{Instance of keyword drop on Bloom-175B}
\label{fig:attacks_5}
\end{figure*}

\begin{figure*}[!ht]
% \minipage{0.49\columnwidth}
% \begin{minted}
% [frame=single,
% framesep=1mm, 
% fontsize=\scriptsize,
% breaklines,
% escapeinside=||,
% highlightlines={7,8,9,10,11,12,13,14,15,16,17},
% ]
% {python}
% def solve(s):
%     "You are given a string s. If s[i] is a letter, reverse its case from Lower to upper or vise versa, otherwise keep it as it is. If the string contains no letters, reverse the string. The function should return the resulted string."
%     Examples:
%     >>> solve("1234") = "4321"
%     >>> solve("ab") = "AB"
%     >>> solve("#a@C") = "#A@c"
%     if not s:
%         return s
%     if not any(c.isalpha() for c in s):
%         return s[::-1]
%     res = ''
%     for c in s:
%         if c.isalpha():
%             res += c.swapcase()
%         else:
%             res += c
%     return res
% \end{minted}
% \endminipage \hfill
% \minipage{0.49\columnwidth}
% \begin{minted}
% [frame=single,
% framesep=1mm, 
% fontsize=\scriptsize,
% breaklines,
% escapeinside=||,
% highlightlines={7,8,9,10},
% highlightcolor={pink}
% ]
% {python}
% def solve(s):
%     "You are given a string s. If s[i] is a letter, case from lower to upper or vise versa, otherwise keep it as it is. If the string contains no letters, reverse the string. The function should return the resulted string."
%     Examples:
%     >>> solve("1234") = "4321"
%     >>> solve("ab") = "AB"
%     >>> solve("#a@C") = "#A@c"
%     if not s.isalpha():
%         return s[::-1]
%     else:
%         return s.swapcase()
% \end{minted}
% \endminipage
\includegraphics[width=\linewidth]{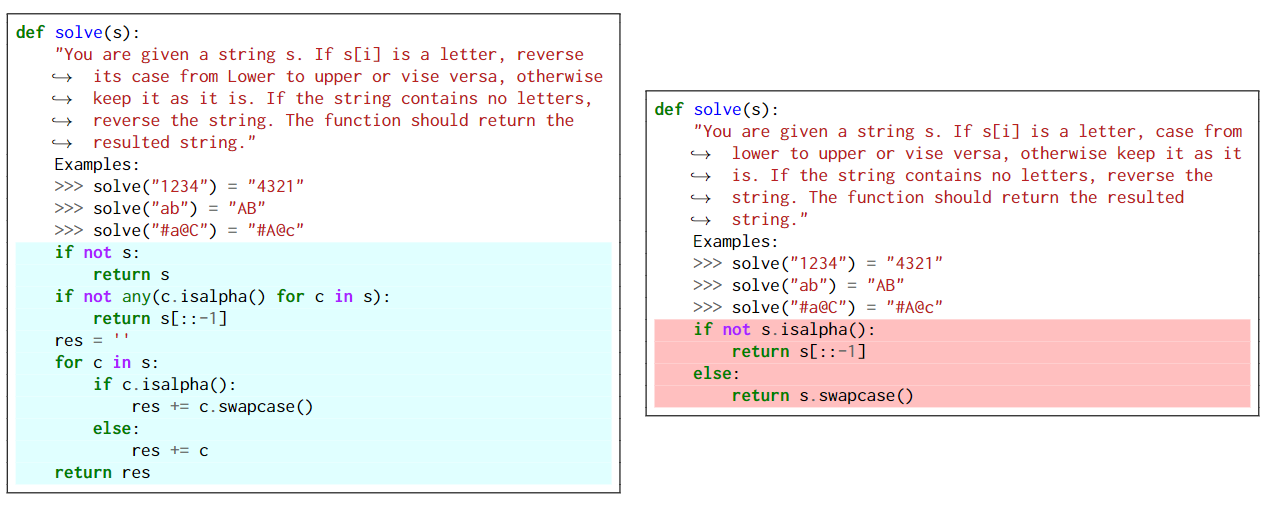}
\caption{Instance of keyword drop on Incoder-1.6B}
\label{fig:attacks_6}
\end{figure*}

\clearpage
\subsection{Quantative Results}
\label{sec:quantapx}
We present our full results table, including the CodeParrot(110M) and Codex(v1) results. Note here that experiments involving the large version of the Bloom Model were done once in the case of pass@100 metric due to restrictions with the API request limits.

\begin{table*}[!htbp]
\begin{center}
\small
\resizebox{0.99\columnwidth}{!}{%
\begin{tabular}{c p{2.5cm} p{2cm} p{2cm} p{2cm} p{2cm}}
\toprule
\multicolumn{1}{c}{} & \multicolumn{1}{c}{} & \multicolumn{2}{c}{Human Eval} & \multicolumn{2}{c}{MBPP}\\
\cmidrule(lr){3-4} \cmidrule(lr){5-6}

\rule{0pt}{2ex} Model & Method\newline of Attack & Pass@1 (T=0.2) & Pass@100 (T=0.8) & Pass@1 (T=0.2) & Pass@100 (T=0.8) \\
\toprule

Codeparrot (110M)~\cite{tunstall2022natural}  & Original       &      3.8  & 12.7   &  5.1  &  26.2\\  
    & Drop One       &     3.3 ($\pm 0.1$)  & 9.7 ($\pm 0.3$)  &  4.1 ($\pm 0.1$) &  16.3 ($\pm 0.5$)\\
    & Drop All       &     3.1 ($\pm 0.1$) & 7.2 ($\pm 0.5$) &  3.9 ($\pm 0.1$) &   15.7 ($\pm 0.7$)\\
    & Drop Ex       &     3.8 ($\pm 0.0$) & 9.9  ($\pm 0.2$) &  5.0 ($\pm 0.0$) &   18.4 ($\pm 0.3$)\\
    & Anon       &     3.4 ($\pm 0.1$) &  8.7 ($\pm 0.2$) &  4.4 ($\pm 0.1$) &  16.1 ($\pm 0.5$)\\
    & Anon+Drop One       &     3.0  ($\pm 0.1$) & 7.5 ($\pm 0.5$)  &  4.0 ($\pm 0.1$) &  13.6 ($\pm 1.1$)\\
    & Anon+Drop All       &     1.9 ($\pm 0.2$) &  6.9 ($\pm 0.5$) &  3.9 ($\pm 0.2$) &   12.0 ($\pm 1.5$)\\
    & Anon+Drop Ex      &     3.4 ($\pm 0.1$) &  8.7 ($\pm 0.3$)  &   4.3 ($\pm 0.2$) & 16.1 ($\pm 0.8$)\\\cmidrule{1-6}

CodeGen-Mono (350M)~\cite{CodeGen} & Original  &  12.7 &	35.2	& 19.2	&  59.4 \\  

& Drop One  &  7.1($\pm 0.1$) &	31.1($\pm 0.4$) &	10.7 ($\pm 0.1$)	& 42.9 ($\pm 0.7$)  \\

& Drop All  & 5.5($\pm 0.1$) &	24.5($\pm 0.6$)	& 9.4 ($\pm 0.2$)	& 38.5 ($\pm 0.7$)  \\

& Drop Ex   &  8.3($\pm 0.1$) & 	29.8($\pm 0.4$)& 	11.7 ($\pm 0.1$) & 	43.6  ($\pm 0.7$) \\

& Anon     & 6.1($\pm 0.2$)	 & 29.8($\pm 0.5$)	 & 12.6 ($\pm 0.1$)	 & 42.6	  ($\pm 0.8$) \\

& Anon+Drop One     &  4.8($\pm 0.2$)  &	28.4($\pm 0.6$)  &	7.6 ($\pm 0.2$)	  &39.4	 ($\pm 1.3$)\\

&Anon+Drop All  &  3.4($\pm 0.2$)	&  22.1($\pm 0.7$)	 & 6.8 ($\pm 0.3$)	&  35.8	 ($\pm 1.6$)\\

& Anon+Drop Ex  & 5.2($\pm 0.2$)	&29.2($\pm 0.6$)	&7.3 ($\pm 0.2$)	&40.5 ($\pm 1.1$)	\\\cmidrule{1-6}

Codeparrot (1.5B)~\cite{tunstall2022natural}  & Original       &      4.1  & 17.8   & 6.1  &  31.2\\  
    & Drop One       &     3.9  ($\pm 0.1$) &  13.2 ($\pm 0.4$) &  4.2 ($\pm 0.2$) &  26.8 ($\pm 0.8$)\\
    & Drop All       &     3.6 ($\pm 0.3$) &  11.1 ($\pm 0.6$) &  3.9 ($\pm 0.2$)&  21.7 ($\pm 1.1$)\\
    & Drop Ex       &     3.7  ($\pm 0.0$) & 14.3 ($\pm 0.2$)  &  5.3 ($\pm 0.0$) &   27.5 ($\pm 0.7$)\\
    & Anon       &     3.8 ($\pm 0.1$) & 12.5 ($\pm 0.2$)  &  4.7 ($\pm 0.2$) &  23.2 ($\pm 0.9$)\\
    & Anon+Drop One       &     3.3 ($\pm 0.2$) &  9.5 ($\pm 0.7$) &  3.9 ($\pm 0.1$) &   20.2 ($\pm 1.5$)\\
    & Anon+Drop All        &     2.1  ($\pm 0.3$)& 8.9 ($\pm 1.1$) &  3.9 ($\pm 0.2$) &   17.9 ($\pm 1.8$)\\
    & Anon+Drop Ex      &     3.7 ($\pm 0.2$) & 11.8 ($\pm 0.9$)  &   4.6 ($\pm 0.1$) &  22.8 ($\pm 0.9$)\\\cmidrule{1-6}

Bloom (1.7B)~\cite{tunstall2022natural}  & Original       &  4.3      &  14.6  & 6.6  & 37.2 \\  
    & Drop One       &  3.0 ($\pm 0.2$)   & 12.2 ($\pm 0.6$) & 2.7 ($\pm 0.3$) & 27.6 ($\pm 1.2$)\\
    & Drop All       &   2.4 ($\pm 0.3$)  & 9.8 ($\pm 0.9$) & 2.6 ($\pm 0.3$) &  24.2 ($\pm 1.8$)\\
    & Drop Ex       &  3.6 ($\pm 0.1$)    & 12.8  ($\pm 0.5$)& 3.1 ($\pm 0.2$) &  29.0 ($\pm 0.9$)\\
    & Anon       &  3.6 ($\pm 0.1$)  & 11.6 ($\pm 0.5$)   & 3.1  ($\pm 0.1$)&  27.5 ($\pm 1.1$)\\
    & Anon+Drop One       &  2.4 ($\pm 0.3$)   & 9.1 ($\pm 1.1$)  & 2.4 ($\pm 0.5$)& 25.3 ($\pm 1.8$)\\
    & Anon+Drop All        &  1.8 ($\pm 0.5$)   & 8.5($\pm 1.3$) & 2.0 ($\pm 0.6$) &   23.1 ($\pm 2.3$)
    \\
    & Anon+Drop Ex      & 3.4 ($\pm 0.2$)   &  11.6 ($\pm 0.6$) &  3.0 ($\pm 0.3$) & 26.7  ($\pm 1.3$)\\\cmidrule{1-6}

Incoder (1.6B)~\cite{fried2022incoder} & Original       &   11.3     &  24.2  &  14.6   &  56.7\\  
    & Drop One       &    10.5 ($\pm 0.1$)  &  22.3 ($\pm 0.9$)  &  11.5($\pm 0.4$)&  45.4 ($\pm 1.1$)\\
    & Drop All       &     9.7 ($\pm 0.3$)  & 17.6 ($\pm 1.2$)   &  12.8 ($\pm 0.6$)&  42.1 ($\pm 1.9$)\\
    & Drop Ex       &     11.3 ($\pm 0.2$)  &  22.2 ($\pm 1.5$)  &  14.4($\pm 0.3$) &  43.8 ($\pm 0.7$)\\
    & Anon       &     9.1 ($\pm 0.1$)  & 21.8 ($\pm 0.8$)  &  11.3 ($\pm 0.5$)  &   45.2 ($\pm 0.8$)\\
    & Anon+Drop One       &     7.4 ($\pm 0.7$)  & 21.5 ($\pm 1.8$)  &  10.5 ($\pm 0.6$)&   44.9 ($\pm 2.4$)\\
    & Anon+Drop All        &     6.3 ($\pm 0.9$)  & 17.5 ($\pm 2.2$)  &  8.0 ($\pm 0.8$)&   41.3($\pm 2.5$)\\
    & Anon+Drop Ex      &    8.7 ($\pm 0.5$)  & 21.3 ($\pm 1.6$)   &  11.2 ($\pm 0.5$)&  43.5($\pm 1.0$)\\

\bottomrule
\end{tabular}
}
\end{center}
\caption{First part of results on Human Eval and MBPP  datasets, for four tested models.}
\label{qual_table1_}
\end{table*}

\begin{table*}[!htbp]
\begin{center}
\small
\resizebox{0.99\columnwidth}{!}{%
\begin{tabular}{c p{2.5cm} p{2cm} p{2cm} p{2cm} p{2cm}}
\toprule
\multicolumn{1}{c}{} & \multicolumn{1}{c}{} & \multicolumn{2}{c}{Human Eval} & \multicolumn{2}{c}{MBPP}\\
\cmidrule(lr){3-4} \cmidrule(lr){5-6}

\rule{0pt}{2ex} Model & Method\newline of Attack & Pass@1 (T=0.2) & Pass@100 (T=0.8) & Pass@1 (T=0.2) & Pass@100 (T=0.8) \\
\toprule

Incoder (6B)~\cite{fried2022incoder} & Original       &      15.2  & 47.0  &  19.4  &  65.1\\  
    & Drop One       &    12.1 ($\pm 0.3$)  & 35.3 ($\pm 1.2$)   &  18.9 ($\pm 0.5$)  &   52.6 ($\pm 1.1$)\\
    & Drop All       &     10.2 ($\pm 0.5$) & 28.2 ($\pm 1.4$)   &  15.6  ($\pm 0.5$)  &  47.0 ($\pm 1.9$)\\
    & Drop Ex       &     12.7 ($\pm 0.3$) &  29.5 ($\pm 0.9$)  &  17.4  ($\pm 0.3$)  &   50.3 ($\pm 0.7$)\\
    & Anon       &     11.6 ($\pm 0.2$) &  32.9 ($\pm 0.9$)   &  14.8  ($\pm 0.6$)  &  50.7 ($\pm 0.8$)\\
    & Anon+Drop One       &     8.1 ($\pm 0.7$) & 30.6 ($\pm 1.7$)   &  13.5 ($\pm 0.7$) &  46.7 ($\pm 2.4$)\\
    & Anon+Drop All        &     7.5 ($\pm 1.3$) & 25.2 ($\pm 2.3$)  &  11.2 ($\pm 1.1$) &   38.9 ($\pm 2.5$)\\
    & Anon+Drop Ex      &     11.2 ($\pm 0.4$) & 28.1 ($\pm 1.1$) &   14.5 
   ($\pm 0.5$)  &  50.2 ($\pm 1.0$) \\\cmidrule{1-6}

CodeGen-Mono (6B)~\cite{CodeGen} & Original       & 26.1 	& 65.8	& 42.3	& 77.3 \\  
    & Drop One       &	18.4 ($\pm 0.3$) &	39.3 ($\pm 0.9$) 	& 25.2 ($\pm 0.5$)	& 65.7  ($\pm 1.2$)\\
    & Drop All       &	13.9 ($\pm 0.4$)&	34.8 ($\pm 1.3$) 	& 22.4 ($\pm 0.6$)	& 57.7  ($\pm 1.6$)\\
    & Drop Ex       & 	20.4 ($\pm 0.3$)& 	42.3 ($\pm 1.1$) & 	27.2 ($\pm 0.5$) & 	61.7  ($\pm 1.1$)\\
    & Anon           & 18.2	 ($\pm 0.3$) & 37.3	 ($\pm 1.0$)  &24.0  ($\pm 0.5$)&	65.6  ($\pm 1.3$)\\
    & Anon+Drop One       &12.6	 ($\pm 0.5$) &24.6	($\pm 1.4$)   &15.8	 ($\pm 0.7$) &58.6  ($\pm 2.2$)\\
    & Anon+Drop All        & 11.5 ($\pm 0.8$)	 & 23.1	($\pm 1.9$)  & 14.9	($\pm 0.8$) & 46.3  ($\pm 2.6$)  \\
    & Anon+Drop Ex      &16.0 ($\pm 0.5$)	&28.3 ($\pm 1.6$) &18.2  ($\pm 0.7$)&	60.7  ($\pm 1.8$)\\\cmidrule{1-6}

Codex (v1)~\cite{chen2021evaluating}  & Original       &      39  & 82.9   &  51.7  &   83.4\\  
    & Drop One       &    29.2  ($\pm 0.2$) &  78 ($\pm 1.3$)  & 48.3  ($\pm 0.4$)& 78.7 ($\pm 1.0$) \\
    & Drop All       &  30  ($\pm 0.4$)   & 67.2 ($\pm 1.7$)   & 33.9  ($\pm 0.8$)& 67.3 ($\pm 1.9$)\\
    & Drop Ex       &  32.9  ($\pm 0.1$)   &  73.7 ($\pm 1.1$) & 42.1  ($\pm 0.2$)& 70.1  ($\pm 0.9$)\\
    & Anon       &     35.3 ($\pm 0.1$) &  81.7 ($\pm 1.2$)    &  50.8  ($\pm 0.2$)&  81.5  ($\pm 1.2$)\\
    & Anon+Drop One       &     23.7 ($\pm 0.5$) & 67.0 ($\pm 2.3$)  &  44.1   ($\pm 0.7$) & 67.7 ($\pm 2.6$)\\
    & Anon+Drop All        &     19.5 ($\pm 0.9 $) &  62.1 ($\pm 2.7$)  &   40.7    ($\pm 1.4$)  &  61.4  ($\pm 3.1$)\\
    & Anon+Drop Ex      &   27.4 ($\pm 0.3$)   &  65.2 ($\pm 1.6$) &  36.7    ($\pm 0.3$)  & 67.7 ($\pm 1.5$) \\\cmidrule{1-6}

Codex (v2)~\cite{chen2021evaluating}  & Original       &      49.4  &   91.4 &  60.1  &  86.3 \\  
    & Drop One       &    36.0  ($\pm 0.1$) &  86.2 ($\pm 0.8$)& 56.0  ($\pm 0.3$)& 79.2 ($\pm 1.1$)\\
    & Drop All       &  37.1 ($\pm 0.3$)   & 73.7  ($\pm 1.3$)& 52.1  ($\pm 0.6$)& 69.5 ($\pm 1.8$)\\
    & Drop Ex       &  41.4  ($\pm 0.1$)   & 81.0  ($\pm 1.1$) & 48.8  ($\pm 0.3$)& 70.7 ($\pm 0.9$)\\
    & Anon       &     44.5 ($\pm 0.2$) &  90.4   ($\pm 1.1$)   &  57.9 ($\pm 0.3$) &  81.7 ($\pm 1.0$)\\
    & Anon+Drop One       &     29.8 ($\pm 0.7$) & 74.4  ($\pm 2.1$)&  51.2  ($\pm 1.1$)  &  69.5 ($\pm 2.3$)\\
    & Anon+Drop All       &     24.2 ($\pm 0.8$) & 68.7  ($\pm 2.8$) &   47.2  ($\pm 1.3$)   &  63.8 ($\pm 3.0$)\\
    & Anon+Drop Ex      &   34.1 ($\pm 0.3$)   & 72.5 ($\pm 1.1$)  &  42.6  ($\pm 0.4$)   & 70.5 ($\pm 1.3$)\\\cmidrule{1-6}

Bloom (176B)~\cite{tunstall2022natural}  & Original       &      16.4  &   57.2 &  20.8  & 62.4 \\  
    & Drop One       &    12.8 ($\pm 0.3$)   &  48.6 & 15.8  ($\pm 0.3$)& 51.4 \\
    & Drop All       &  11.5 ($\pm 0.6$)     &  40.2  & 14.2 ($\pm 0.5$)& 44.4 \\
    & Drop Ex       &  15.2  ($\pm 0.2$)    & 43.3  & 15.8 ($\pm 0.2$)& 50.1 \\
    & Anon       &     14.0 ($\pm 0.3$)  &  48.3   &  15.1 ($\pm 0.1$)&  51.2\\
    & Anon+Drop One       &     12.8 ($\pm 0.4$) & 41.9 &  13.6 ($\pm 0.7$)&  46.8\\
    & Anon+Drop All       &     10.3 ($\pm 0.8$)  & 36.8  &  12.6 ($\pm 1.1$)& 38.4\\
    & Anon+Drop Ex      &   14.0 ($\pm 0.3$)    & 39.8  &  14.3 ($\pm 0.3$)& 47.8\\

\bottomrule
\end{tabular}
}
\end{center}
\caption{Second part of results on Human Eval and MBPP datasets, for four tested models.}
\label{qual_table2_}
\end{table*}

\clearpage
\subsection{Few Interesting Examples}
\begin{figure*}[!ht]
% \begin{minted}
% [frame=single,
% framesep=1mm, 
% fontsize=\scriptsize,
% breaklines,
% escapeinside=||,
% highlightlines={9,10,11,12,13,14,15,16,17},
% highlightcolor={pink}
% ]
% {python}
% from typing import List

% def func(numbers: List[float], threshold: float) -> bool:
%     " Check if in given list of numbers,are any two numbers closer to each other than given threshold."
%     >>> func({1.0, 2.0, 3.0], 10.5)
%     False
%     >>> func((1.0, 2.8, 3.0, 4.0, 5.0, 2.0], 0.3)
%     True
%     for i in range(len(numbers) - 1):
%         if abs(numbers[i].id == id) {
%             if (temp.id == id) {
%                 retum temp;
%             }
%         }
%     }
%     return null;
% \end{minted}
\includegraphics[width=\linewidth]{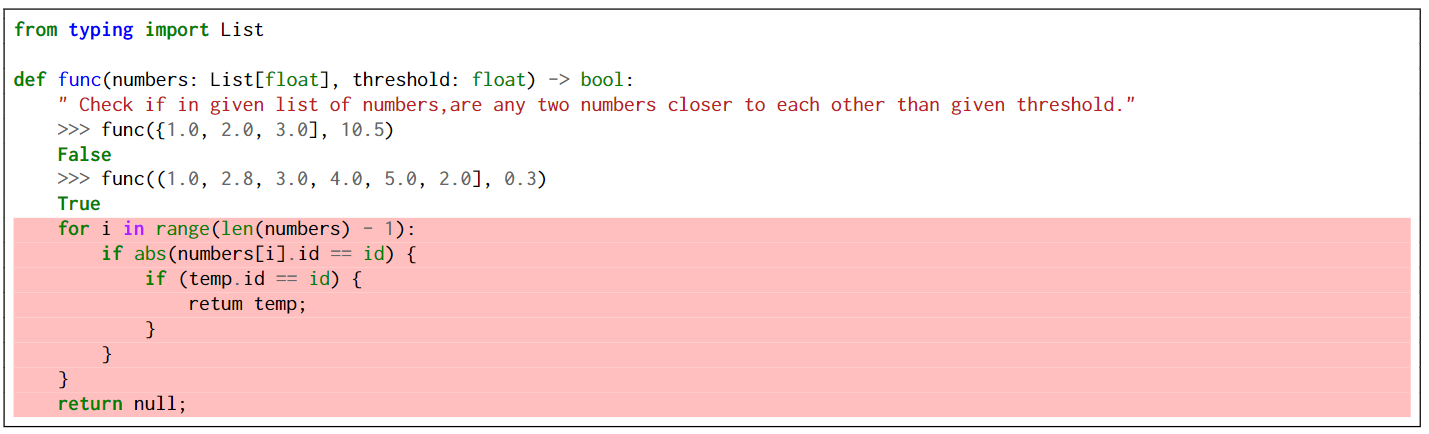}
\caption{Bloom (175B) using Javascript instead of Python3 to complete a function with the \textit{Anonymize} transformation.}
\label{fig:js}
\end{figure*}

\begin{figure*}[!ht]
% \begin{minted}
% [frame=single,
% framesep=1mm, 
% fontsize=\scriptsize,
% breaklines,
% escapeinside=!!,
% highlightlines={4,5,6,7,8,9,10,11,12},
% highlightcolor={pink}
% ]
% {python}
% def func(decimal):
%     "You will be given a decimal form and your task is to convert it to binary format. The function should return a string, with each character representing a binary number. Each character in the string will be '0' or '1'. There will be an extra couple of characters 'db' at the beginning and at the end of the string. The extra characters are there to help with the format."
    
%     </file source=github filename=binary_numbers.py |>
%     <|c|>
%     Why does the db string appear !in! the output!?! And why does 1 get converted to the empty string!?!.
%     My code also leaves out the !0! at the beginning.
%     <|/c|>
%     <| a tags=python-3.x, python |>
%     <code>
%     n = int(input('Give the number you want in binary form'))
%     remainder = n%2
% \end{minted}
\includegraphics[width=\linewidth]{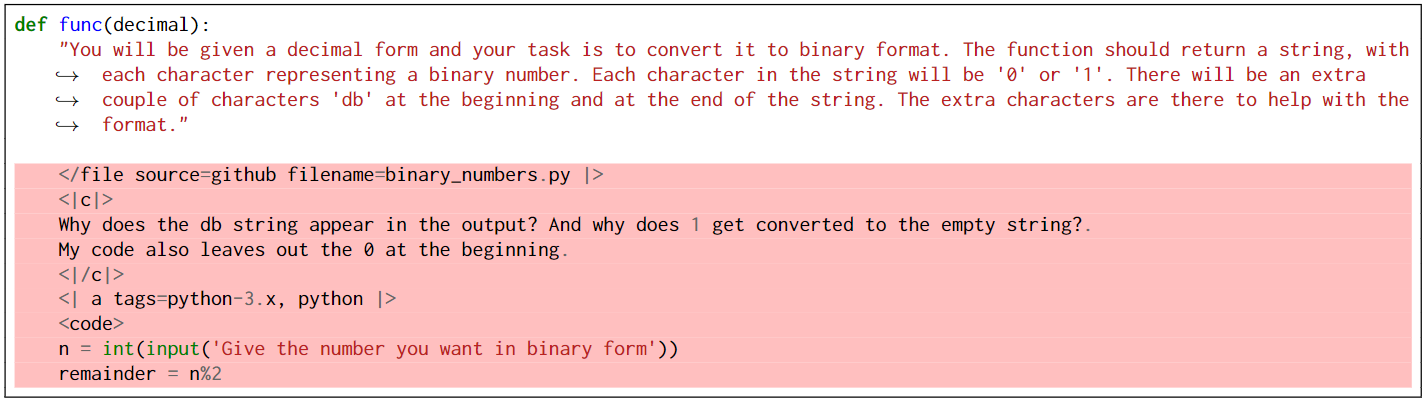}
\caption{Incoder (6B) disclosing the name of a file as well as some human-like questions when faced with a \textit{Anonymize + Drop One} transformation.}
\label{fig:leak_1}
\end{figure*}

\begin{figure*}[!ht]
\includegraphics[width=\linewidth]{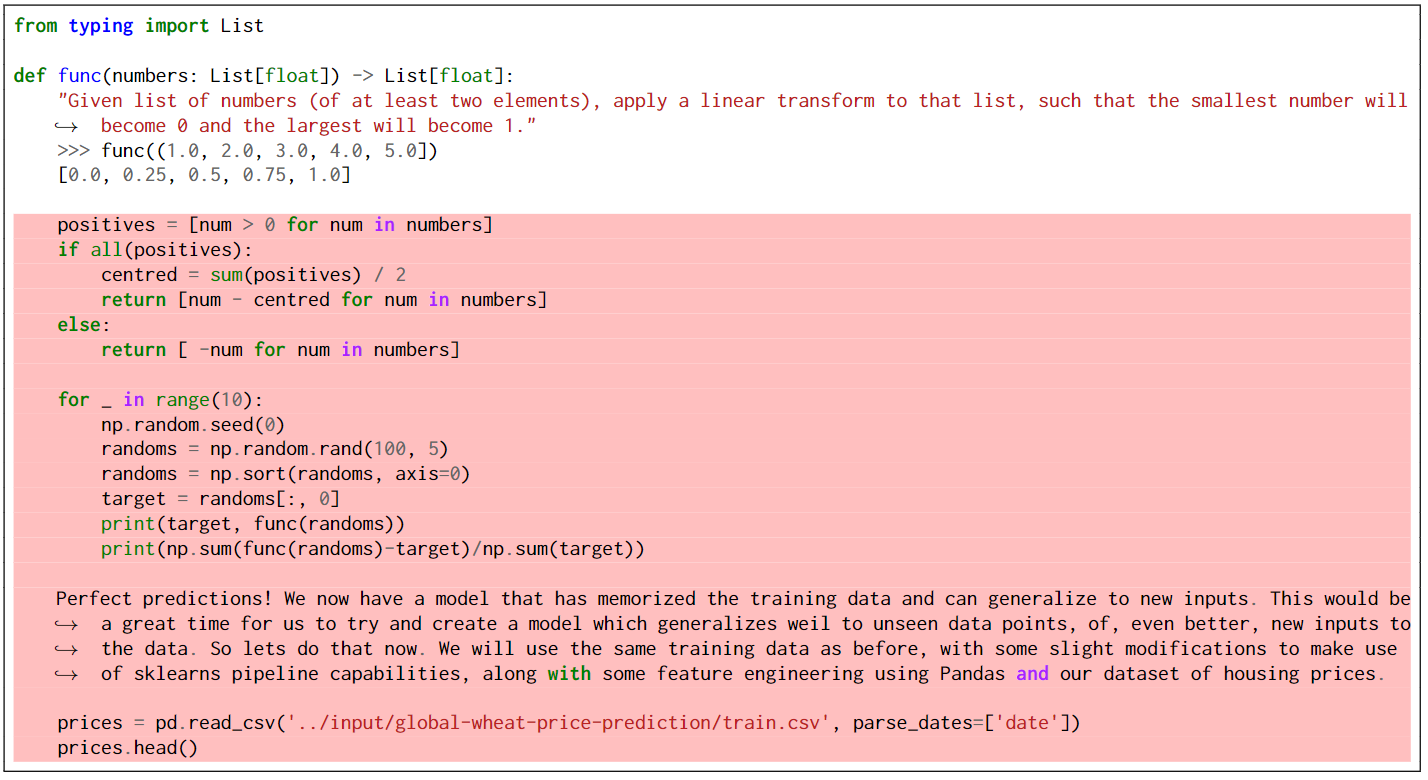}
\caption{Incoder (1.6B) adding some snippet of ambiguous functionality followed by something that looks like some exercise comments.}
\label{fig:leak_2}
\end{figure*}

\begin{figure*}[!ht]
\includegraphics[width=\linewidth]{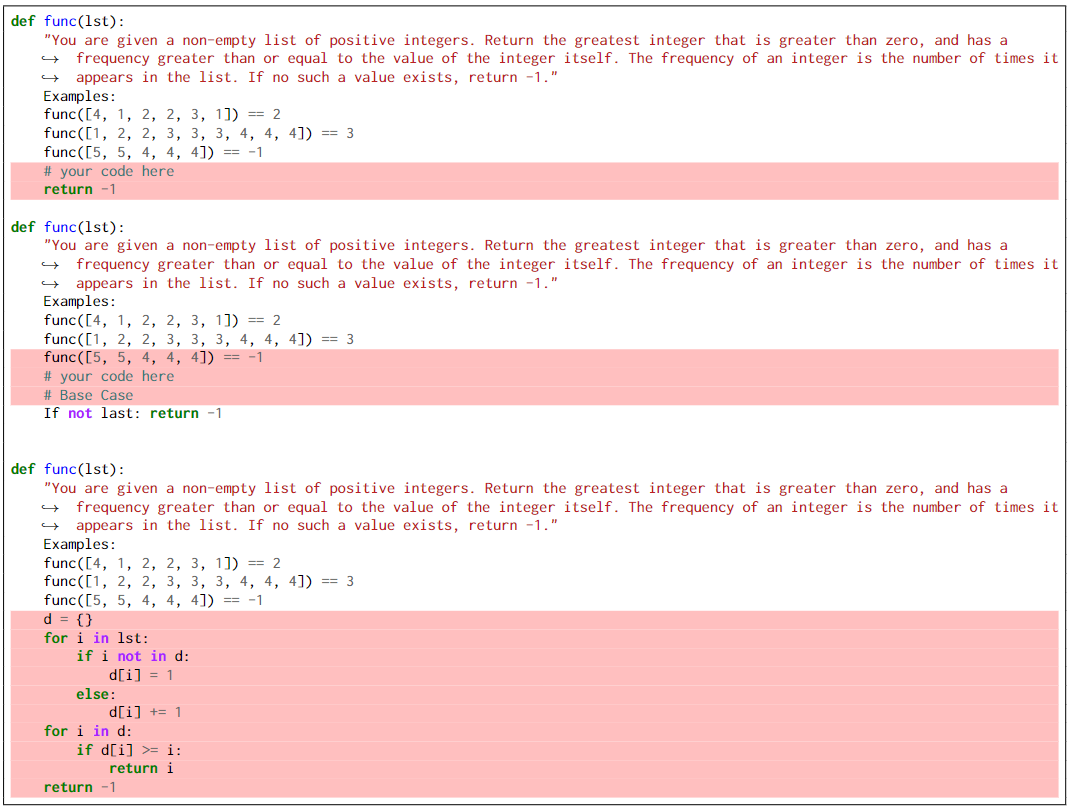}
\caption{Three different faulty instances of Codex (v1) completions to an anonymized problem.}
\label{fig:sys_1}
\end{figure*}

\clearpage
\subsection{Algorithms}
\label{sec:algorithms}
\begin{algorithm}[!ht]
\caption{Block of Influence Splitting}\label{alg:bis}
\begin{algorithmic}[1]

\STATE {$cc:$ Code Challenge Instance}
\newline \textcolor{teal}{\# Locate function name, which is the next token after the last matched "def", and keep start and end index of it.}
\STATE {$name, start\_name\_index, end\_name\_index \gets NameMatch(cc)$} 
\newline \textcolor{teal}{\# Anything prior to the match, such as imports or helper functions is considered prefix.}
\STATE {$prefix \gets cc[:start\_name\_index]$} 
\newline \textcolor{teal}{\# Look for tokens such as (Example, example, $>$, $\gg$). If no matches were found, look for uses of the function name in the challenge.}
\IF {$ExampleMatch(cc[end\_name\_index:]) \neq None$}
\STATE {$examples, start\_example\_index \gets ExampleMatch(cc[end\_name\_index:])$}
\ELSE
\STATE {$examples, start\_example\_index \gets FunctionMatch(cc[end\_name\_index:])$}
\ENDIF
\textcolor{teal}{\newline\# The description should fall between the function name and the examples.}
\STATE {$description \gets cc[end\_name\_index:start\_example\_index]$}
\textcolor{teal}{\newline \# Form the blocks and return.}
\STATE {$Name Block \gets prefix + name$}
\STATE {$Description Block \gets description$}
\STATE {$Example Block \gets examples$}
\end{algorithmic}
\end{algorithm}

\begin{algorithm}[!ht]
\caption{Keyword Identification}\label{alg:ki}
\begin{algorithmic}[1]

\STATE {$KB:$ The KeyBert model} 
\STATE {$nb:$ Name Block}
\STATE {$db:$ Description Block}
\STATE {$eb:$ Example Block}
\STATE {$kw: \gets \emptyset$ Keywords}
\STATE {$fkw: \gets \emptyset$ Filtered Keywords}
\textcolor{teal}{\newline\# Use the model to extract some initial unigram and bigram keywords.}
\STATE {$kw \gets KB(db)$}
\textcolor{teal}{\newline\# Filter out keywords non-related to coding.}
\FOR{$i$ $in$ $kw$}
    \IF {$cossim(i, [Python, Programming, Code]) > 0.7$}
        \IF {$stem(i) \in [nb,eb]$ $or$ $equiv(i) \in [nb,eb]$}
            \STATE {$fkw \gets i$}
        \ENDIF
    \ENDIF
\ENDFOR
\STATE \textbf{return}
\end{algorithmic}
\end{algorithm}

\begin{algorithm}[!ht]
\caption{Transformation and Execution}\label{alg:te}
\begin{algorithmic}[1]

\STATE {$CM:$ The code generation model}
\STATE {$cc:$ A coding challenge instance}
\STATE {$nb:$ Name Block}
\STATE {$fkw:$ Filtered Keywords}
\STATE {$db:$ Description Block}
\STATE {$eb:$ Example Block}
\STATE {$org\_pa1:$ Original Pass@1 score}
\STATE {$tra\_pa1:$ Transformed Pass@1 score}
\STATE {$org\_pa100:$ Original Pass@100 score}
\STATE {$tra\_pa100:$ Transformed Pass@1 score}
\STATE {$mode:$ The transformation mode}
\textcolor{teal}{\newline\# Measure initial performance on the challenge}

\STATE {$org\_pa1,org\_pa100  \gets CM(cc, T=0.2), CM(cc, T=0.8)$}

\IF {$mode = 0$}
\STATE {$cc\_new \gets swap(nb, "func") + db + eb$}\textcolor{teal}{\ \# Anonymization}
\ELSIF {$mode = 1$}
\STATE {$cc\_new \gets nb + remove\_kw(db, choose\_single(fkw)) + eb$}\textcolor{teal}{\ \# Drop One}
\ELSIF {$mode = 2$}
\STATE {$cc\_new \gets nb + remove\_kw(db, fkw) + eb$}\textcolor{teal}{\ \# Drop All}
\ELSIF {$mode = 3$}
\STATE {$cc\_new \gets nb + db$}\textcolor{teal}{\ \# Drop Examples}
\ELSIF {$mode = 4$}
\STATE {$cc\_new \gets swap(nb, "func") + remove\_kw(db, choose\_single(fkw)) + eb$}\textcolor{teal}{\ \# Anonymization + Drop One}
\ELSIF {$mode = 5$}
\STATE {$cc\_new \gets swap(nb, "func") + remove\_kw(db, fkw) + eb$}\textcolor{teal}{\ \# Anonymization + Drop All}
\ELSIF {$mode = 6$}
\STATE {$cc\_new \gets swap(nb, "func") + db$}\textcolor{teal}{\ \# Anonymization + Drop Examples}
\ENDIF
\STATE {$tra\_pa1,tra\_pa100  \gets CM(cc\_new, T=0.2), CM(cc\_new, T=0.8)$}
\STATE {$dif\_1 \gets \frac{tra\_pa1 - org\_pa1}{tra\_pa1}$}
\STATE {$dif\_100 \gets \frac{tra\_pa100 - org\_pa100}{tra\_pa100}$}
\STATE \textbf{return dif\_1, dif\_100} 
\end{algorithmic}
\end{algorithm}

\clearpage
\subsection{On the effect of function names}
\label{func_name}

Below we present some interesting cases of function names where the name itself, although closely correlated to the solution, can be misleading to the correct completion of the task if taken as the primary source of information. 
We provide our intuitions and completions from Codex-v2 when asked to generate a function based only on its name.

\begin{itemize}
    \item \textbf{Name}: sort\_numbers \\
    \textbf{Description}: Input is a space-delimited string of numerals from 'zero' to 'nine.' Valid choices are 'zero,' 'one,' 'two,' 'three,' 'four,' 'five,' 'six,' 'seven,' 'eight,' and 'nine.' Return the string with numbers sorted from smallest to largest.\\
    \textbf{Comment}: Looking only at sort\_numbers, a typical response would be to write a common sorting algorithm.\\
    \textbf{Most common Codex completions @ (T=0.2, p=0.95) and @ (T=0.6, p=0.95)}:
    \begin{figure*}[!ht]
    \centering
        \includegraphics[width=0.5\linewidth]{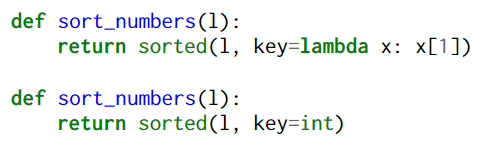}
    \end{figure*}
    
    % \begin{minted}[mathescape, fontsize=\small]{python}
    % def sort_numbers(l):
    %     return sorted(l, key=lambda x: x[1])
        
    % def sort_numbers(l):
    %     return sorted(l, key=int)
    % \end{minted}

    \item \textbf{Name}: below\_zero \\
    \textbf{Description}: You're given a list of deposit and withdrawal operations on a bank account that starts with a zero balance. Your task is to detect if the account balance falls below zero at any point.\\
    \textbf{Comment}: Looking only at below\_zero, a typical response would be to write a logical check of an input number with zero.\\
    \textbf{Most common Codex completions @ (T=0.2, p=0.95) and @ (T=0.6, p=0.95)}:
    \begin{figure*}[!ht]
    \centering
        \includegraphics[width=0.25\linewidth]{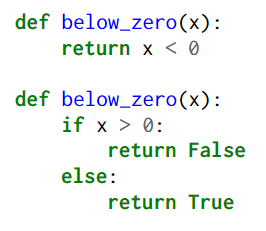}
    \end{figure*}
    % \begin{minted}[mathescape, fontsize=\small]{python}
    % def below_zero(x):
    %     return x < 0
        
    % def below_zero(x):
    %     if x > 0:
    %         return False
    %     else:
    %         return True
    % \end{minted}
    \newpage
    \item \textbf{Name}: x\_or\_y \\
    \textbf{Description}: A simple program that should return the value of x if n is a prime number and should return the value of y otherwise.\\
    \textbf{Comment}: Looking only at x\_or\_y, a typical response would be to write a logical check of $x \land y$.\\
    \textbf{Most common Codex completions @ (T=0.2, p=0.95) and @ (T=0.6, p=0.95)}:
    \begin{figure*}[!ht]
    \centering
        \includegraphics[width=0.25\linewidth]{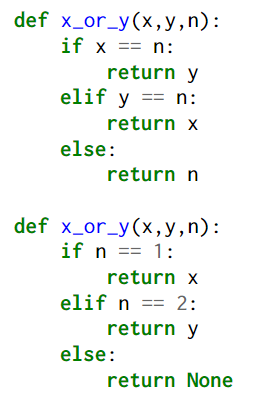}
    \end{figure*}
    % \begin{minted}[mathescape, fontsize=\small]{python}
    % def x_or_y(x,y,n):
    %     if x == n:
    %         return y
    %     elif y == n:
    %         return x
    %     else:
    %         return n

    % def x_or_y(x,y,n):
    %     if n == 1:
    %         return x
    %     elif n == 2:
    %         return y
    %     else:
    %         return None
    % \end{minted}
    \item \textbf{Name}: find\_max \\
    \textbf{Description}: Write a function that accepts a list of strings. The list contains different words. Return the word with the maximum number of unique characters. If multiple strings have the maximum number of unique characters, return the one which comes first in lexicographical order.\\
    \textbf{Comment}: Looking only at find\_max, a typical response would be to write a function that returns the max element from an iterable.\\
    \textbf{Most common Codex completions @ (T=0.2, p=0.95) and @ (T=0.6, p=0.95)}:
    \begin{figure*}[!ht]
    \centering
        \includegraphics[width=0.3\linewidth]{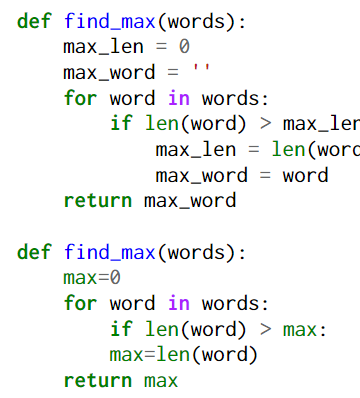}
    \end{figure*}
    % \begin{minted}[mathescape, fontsize=\small]{python}
    % def find_max(words):
    %     max_len = 0
    %     max_word = ''
    %     for word in words:
    %         if len(word) > max_len:
    %             max_len = len(word)
    %             max_word = word
    %     return max_word
        
    % def find_max(words):
    %     max=0
    %     for word in words:
    %         if len(word) > max:
    %         max=len(word)
    %     return max
    % \end{minted}
\end{itemize}
\clearpage
\subsection{Looking for copying patterns in open-source dataset}
\label{nearperfectcopies}
One of the major benefits of including the CodeParrot models in our analysis is their open-source dataset. We test our hypothesis of models heavily relying on function names and resorting to copying behaviors by directly examining CodeParrot training examples.
Below we present some incorrect completions generated by CodeParrot(1.5B) and the "heavily inspired" files where the responses were found. Note that in the case of anonymization, responses differ.

\begin{itemize}
\item  \textbf{1 - CodeParrot Completion}
% \begin{minted}[mathescape,breaklines, fontsize=\small]{python}
% def generate_random_string(s):
%     "Write a function that generates a random string using the seed s."
%     return " ".join(random.sample(string.ascii_letters + string.digits,s))
% \end{minted}
    \begin{figure*}[!ht]
    \centering
        \includegraphics[width=0.75\linewidth]{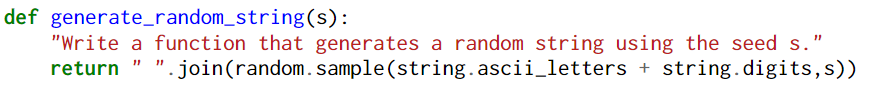}
    \end{figure*}

\textbf{This is found in:}\\
'blockly/appengine/storage.py'\\
'Android/platforms/android/assets/www/Blockly/appengine/storage.py'\\

\textbf{CodeParrot Completion Anonymized}
% \begin{minted}[mathescape, breaklines, fontsize=\small]{python}
% def func(s):
%     "Write a function that generates a random string using the seed s."
%     return random.choice(s)
% \end{minted}
    \begin{figure*}[!ht]
    \centering
        \includegraphics[width=0.75\linewidth]{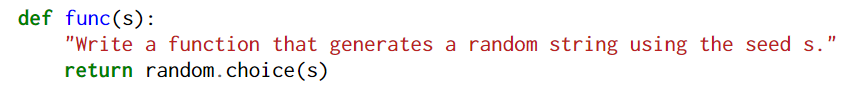}
    \end{figure*}

\item  \textbf{2 - CodeParrot Completion}
% \begin{minted}[mathescape, breaklines, fontsize=\small]{python}
% def perimeter(s):
%     "Return the perimeter of a square with side length s."
%     return perimeter(s).magnitude()
% \end{minted}
    \begin{figure*}[!ht]
    \centering
        \includegraphics[width=0.75\linewidth]{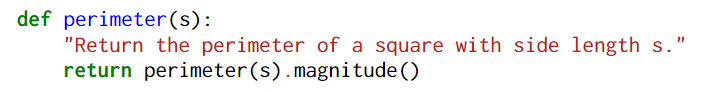}
    \end{figure*}

\textbf{This is found in:}\\
‘bretglun/fwqpbo/fatWaterSeparation.py’\\
'indico/numpy/distutils/tests/test\_misc\_util.py'\\

\textbf{CodeParrot Completion Anonymized}
% \begin{minted}[mathescape,breaklines, fontsize=\small]{python}
% def func(s):
%     "Return the perimeter of a square with side length s."
%     return (s * 0.5)
% \end{minted}
    \begin{figure*}[!ht]
    \centering
        \includegraphics[width=0.75\linewidth]{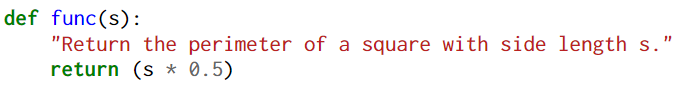}
    \end{figure*}
\clearpage
\item  \textbf{3 - CodeParrot Completion}
% \begin{minted}[mathescape,breaklines, fontsize=\small]{python}
% def tostring(element):
%     "Convert the given element to string representation"
%     return str(element.toxml())
% \end{minted}
    \begin{figure*}[!ht]
    \centering
        \includegraphics[width=0.75\linewidth]{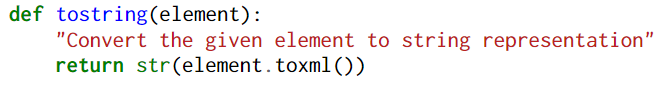}
    \end{figure*}

\textbf{This is found in:}\\
‘warrior/Framework/Utils/xml\_Utils.py’\\

\textbf{CodeParrot Completion Anonymized}
% \begin{minted}[mathescape, breaklines, fontsize=\small]{python}
% def func(s):
%     "Convert the given element to string representation"
%     return str(element)
% \end{minted}
    \begin{figure*}[!ht]
    \centering
        \includegraphics[width=0.75\linewidth]{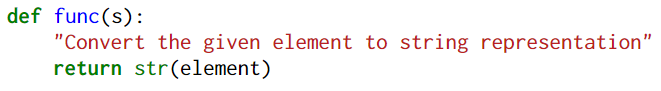}
    \end{figure*}

\item  \textbf{4 - CodeParrot Completion}
% \begin{minted}[mathescape,breaklines, fontsize=\small]{python}
% def add(s, p):
%     "Add s to p and return the result"
%     if not s:
%         return None
%     if not isinstance(p, list):
%         p = [p]
%         p.append(s)
%     return
% \end{minted}
    \begin{figure*}[!ht]
    \centering
        \includegraphics[width=0.55\linewidth]{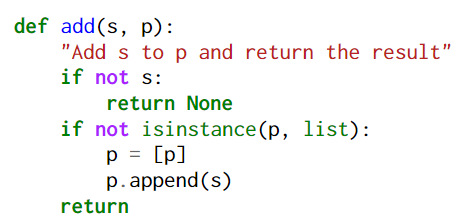}
    \end{figure*}

\textbf{This is found in:}\\
'couchpotato/core/plugins/status/main.py'\\
'modules/wifi/nano-c047.12/WiFiEngine/tools/axfbin.py'\\

\textbf{CodeParrot Completion Anonymized}
% \begin{minted}[mathescape,breaklines, fontsize=\small]{python}
% def func(s):
%     "Add s to p and return the result"
%     return s + p
% \end{minted}
    \begin{figure*}[!ht]
    \centering
        \includegraphics[width=0.55\linewidth]{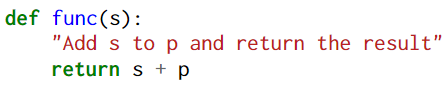}
    \end{figure*}
\clearpage
\item  \textbf{5 - CodeParrot Completion}
% \begin{minted}[mathescape,breaklines, fontsize=\small]{python}
% def count_distinct_characters(string: str):
%   "Given a string, find out how many distinct characters (regardless of case) does it consist of"
%   count = 0
%   for char in string:
%     if char in string:
%         count += 1
%     return count
% \end{minted}
    \begin{figure*}[!ht]
    \centering
        \includegraphics[width=\linewidth]{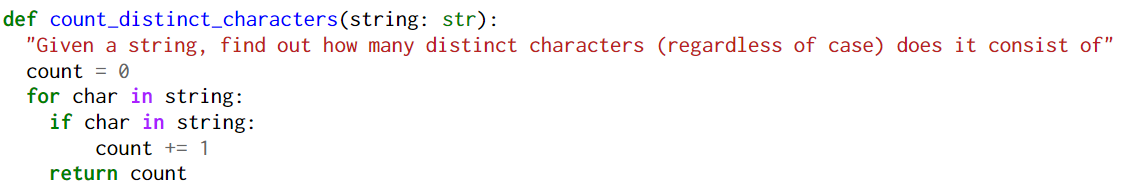}
    \end{figure*}

\textbf{This is found in:}\\
'desktop/core/ext-py/guppy-0.1.10/guppy/heapy/Prof.py'\\
'cpp/scons/scons-local-2.0.0.final.0/SCons/Util.py'\\

\textbf{CodeParrot Completion Anonymized}
% \begin{minted}[mathescape,breaklines, fontsize=\small]{python}
% def func(string: str):
%   "Given a string, find out how many distinct characters (regardless of case) does it consist of"
%     return len(re.findall(r"[^a-zA-Z0-9]", string))

% \end{minted}
    \begin{figure*}[!ht]
    \centering
        \includegraphics[width=\linewidth]{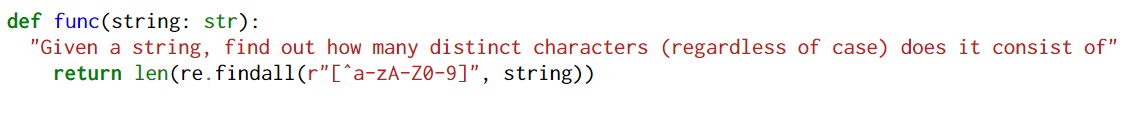}
    \end{figure*}
\end{itemize}
\clearpage
\subsection{Attention View}
In this section, we present illustrations of attention patterns. We use Codeparrot (330M) as our target model, before and after the combined finetuning process and create visualizations for two coding challenges. The first challenge is:
% \begin{minted}[mathescape, breaklines, fontsize=\small]{python}
% def tostring(element):
%     "Convert the given element to string representation"
%      Examples:
%      >>> tostring(1)
%      "1"
%      >>> tostring("obj")
%      "obj"
% \end{minted}
    \begin{figure*}[!ht]
    \centering
        \includegraphics[width=0.75\linewidth]{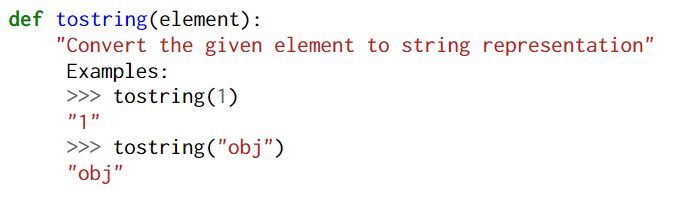}
    \end{figure*}

and the second challenge is:

% \begin{minted}[mathescape, breaklines, fontsize=\small]{python}
% import math

% def perimeter(s):
%     "Return the perimeter of a square with side length s."
%     Examples:
%     >>> perimeter(1)
%     1
%     >>> perimeter(math.sqrt(2))
%     2
% \end{minted}
    \begin{figure*}[!ht]
    \centering
        \includegraphics[width=0.75\linewidth]{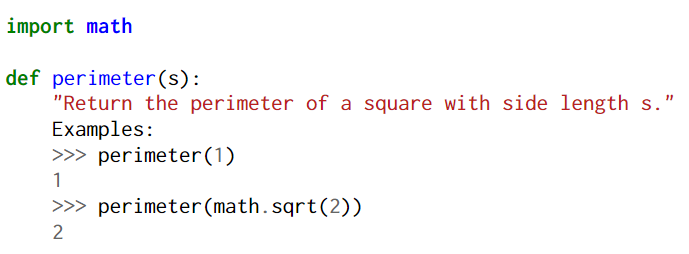}
    \end{figure*}

\begin{figure}[!ht]%
    \centering
    \subfloat[\centering ]{{\includegraphics[width=0.65\linewidth]{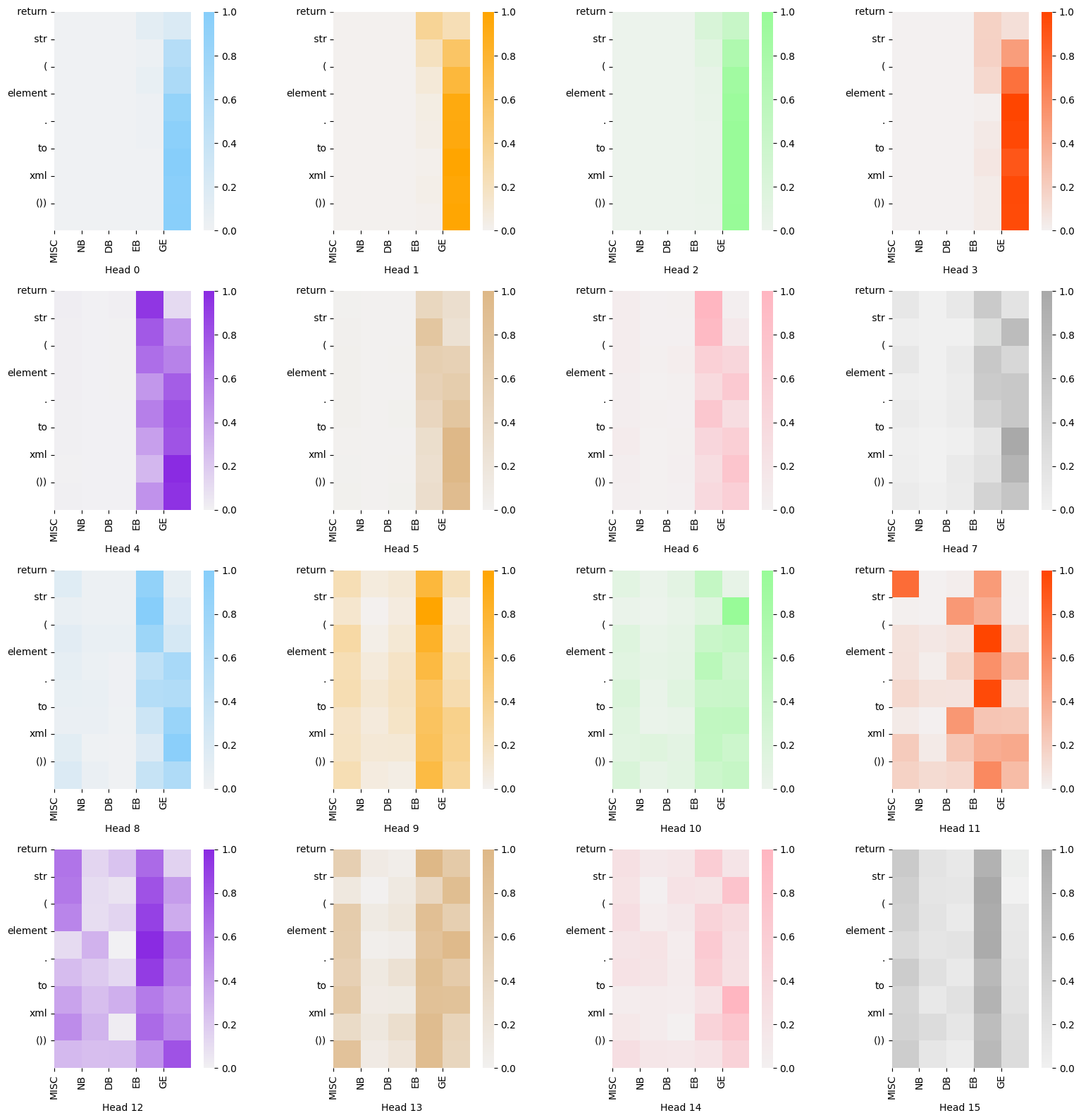} }}
    \hspace{0.05\linewidth}
    \subfloat[\centering ]{{\includegraphics[width=0.65\linewidth]{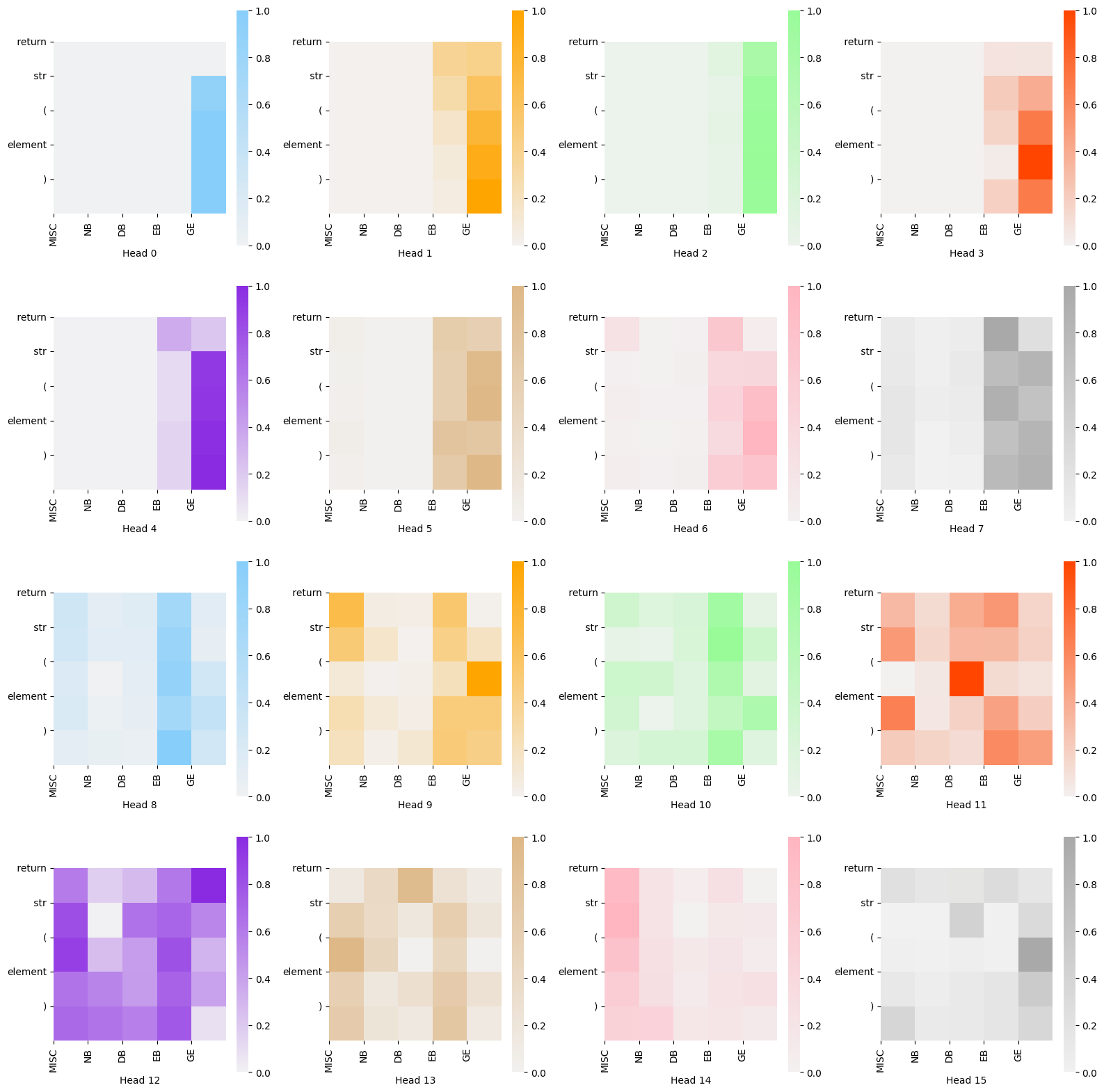} }}%
    \caption{\small{Illustrated attention scores of "tostring" coding challenge before (a) and after (b) augmentation (Layer 4).}}%
    \label{fig:before_1}
\end{figure}

\begin{figure}[!ht]%
    \centering
    \subfloat[\centering ]{{\includegraphics[width=0.65\linewidth]{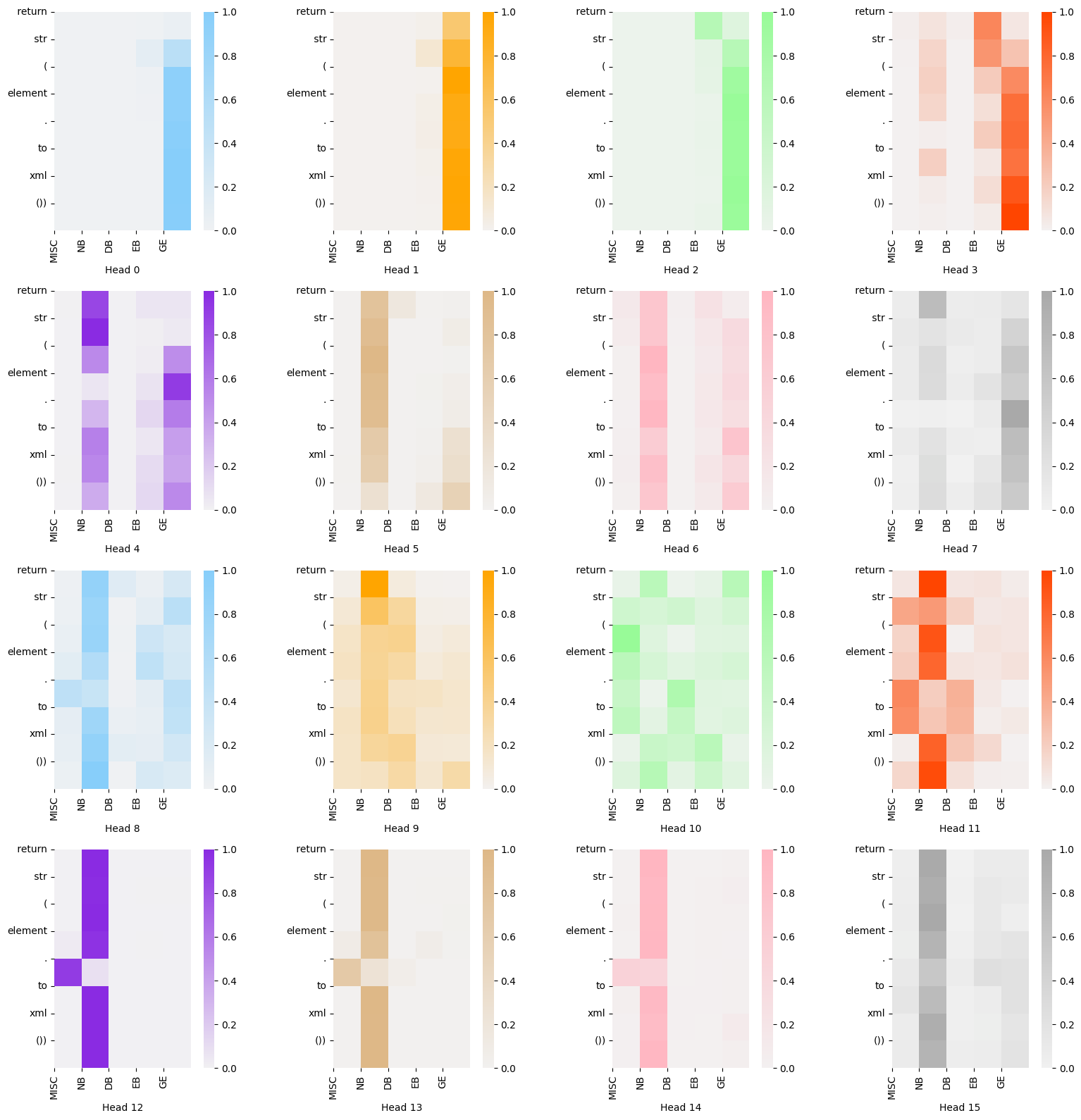} }}
    \hspace{0.05\linewidth}
    \subfloat[\centering ]{{\includegraphics[width=0.65\linewidth]{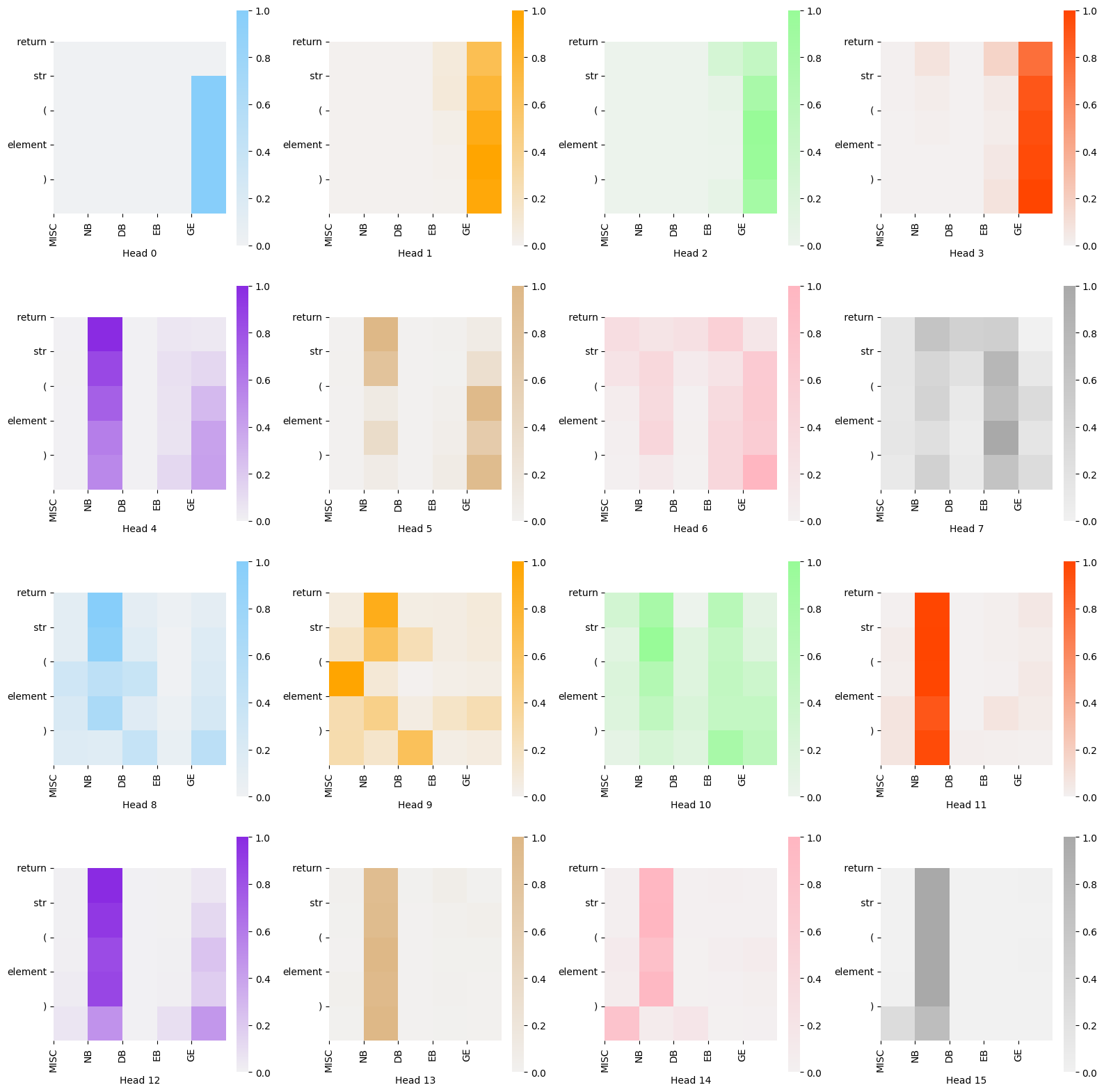} }}
    \caption{\small{Illustrated attention scores of "tostring" coding challenge before (a) and after (b) augmentation (Layer 8).}}%
    \label{fig:after_1}
\end{figure}

\begin{figure}[!ht]%
    \centering
    \subfloat[\centering ]{{\includegraphics[width=0.65\linewidth]{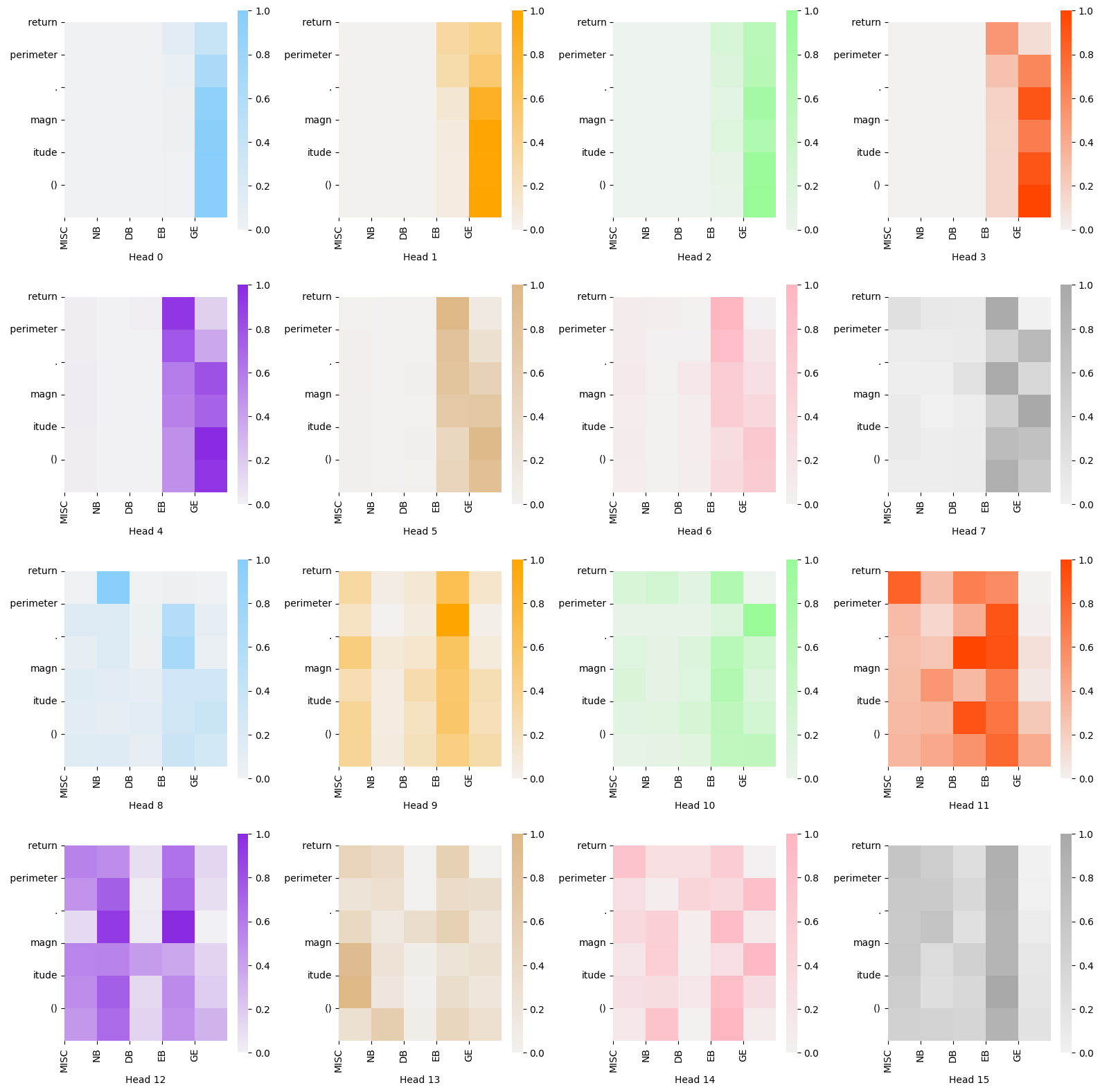} }}
    \hspace{0.05\linewidth}
    \subfloat[\centering ]{{\includegraphics[width=0.65\linewidth]{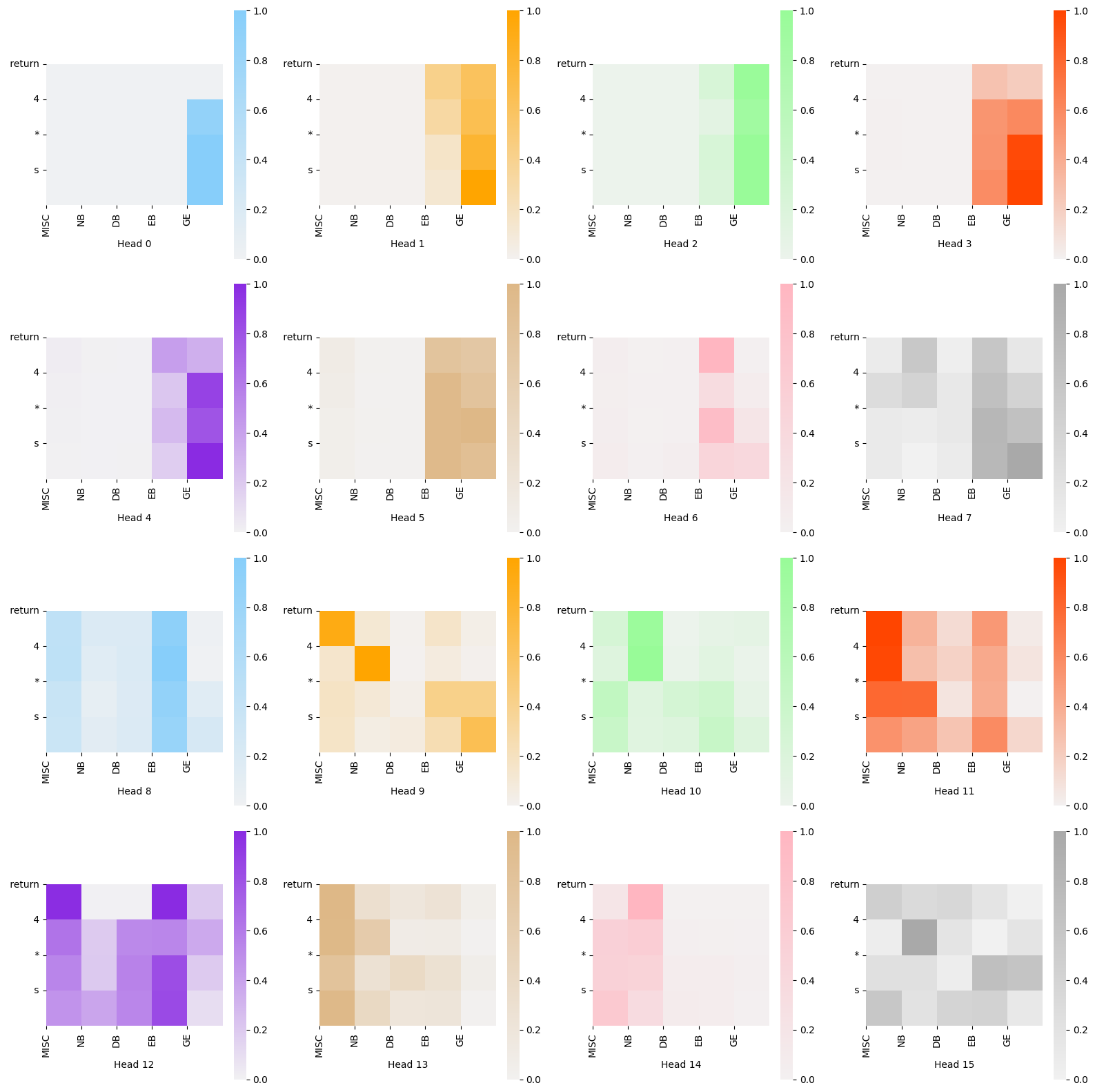} }}
    \caption{\small{Illustrated attention scores of "perimeter" coding challenge before (a) and after (b) augmentation (Layer 4).}}%
    \label{fig:before_2}
\end{figure}

\begin{figure}[!ht]%
    \centering
    \subfloat[\centering ]{{\includegraphics[width=0.65\linewidth]{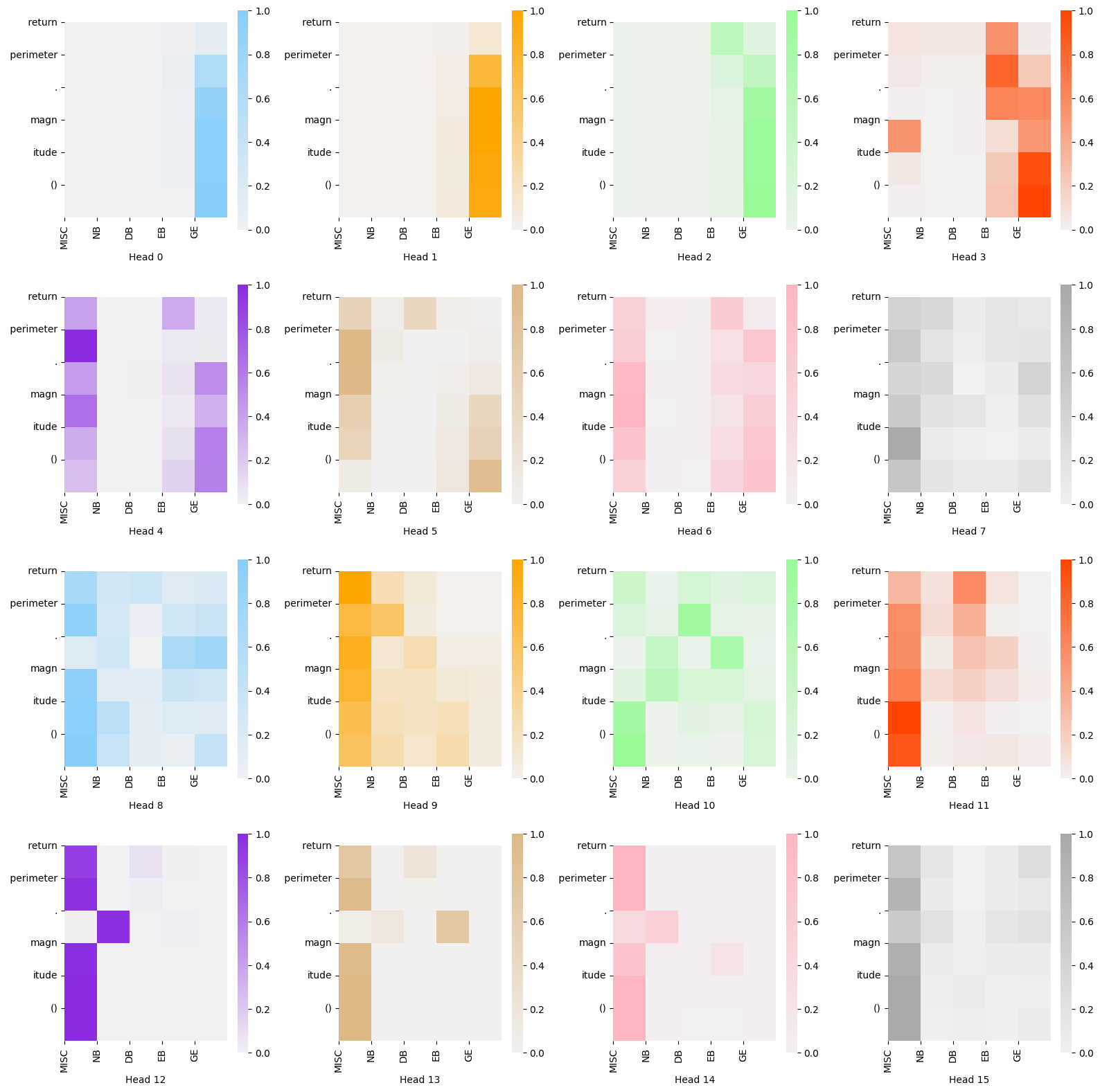} }}
    \hspace{0.05\linewidth}
    \subfloat[\centering ]{{\includegraphics[width=0.65\linewidth]{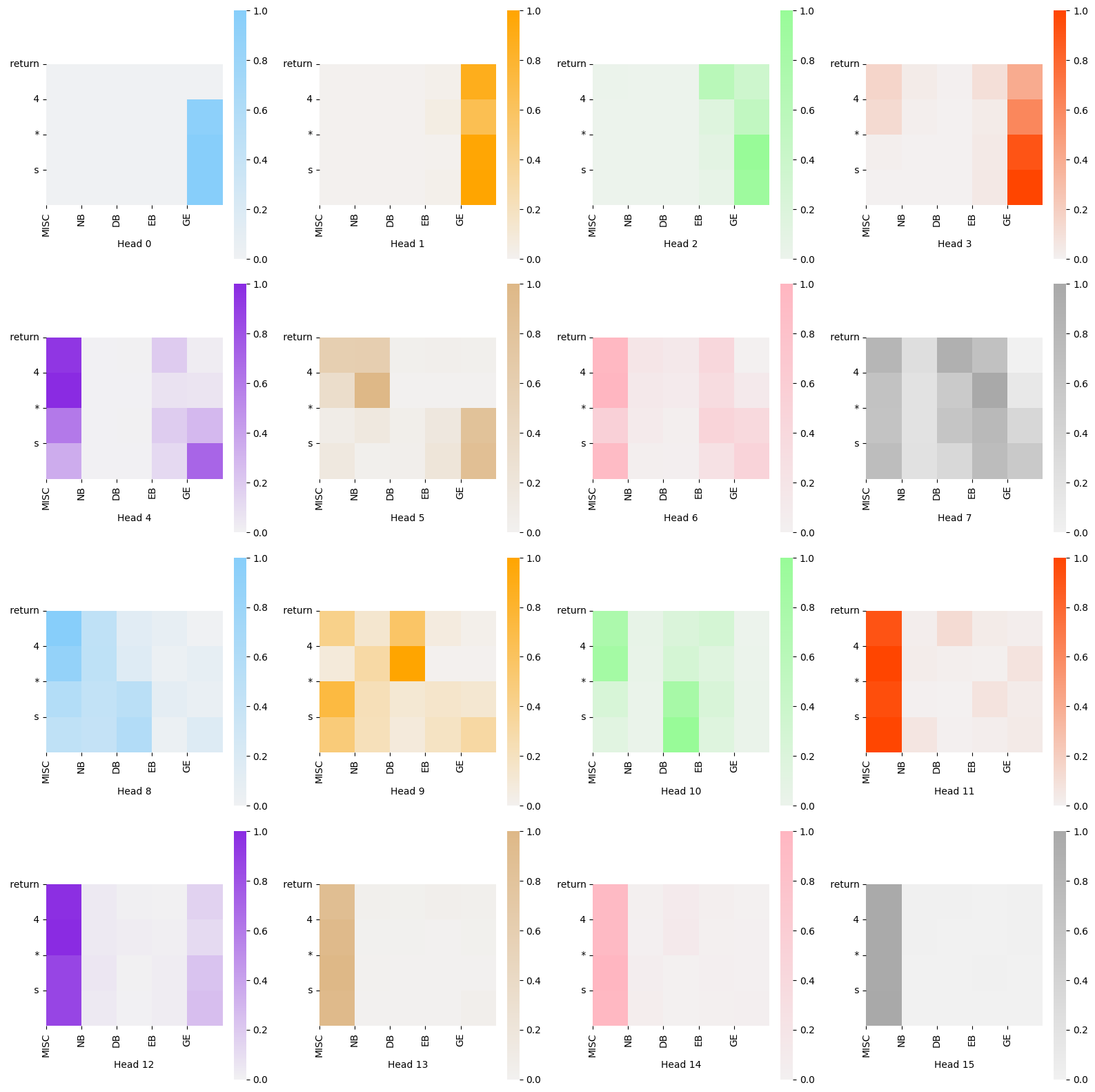} }}
    \caption{\small{Illustrated attention scores of "perimeter" coding challenge before (a) and after (b) augmentation (Layer 8).}}%
    \label{fig:after_2}
\end{figure}

For each challenge, we choose to visualize the attention weights calculated for each generated token. We group together tokens of each challenge into five categories:
\begin{itemize}
    \item \textbf{NB}: All tokens belonging to the \nb
    \item \textbf{DB}: All tokens belonging to the \db
    \item \textbf{EB}: All tokens belonging to the \eb
    \item \textbf{GE}: The so-far model generated tokens (solution)
    \item \textbf{MISC}: Any remaining tokens such as prefixes and imports.
\end{itemize}
Our goal is to detect whether augmentations can cause visible changes to the attention patterns over the \textit{Blocks of Influence}.
In our analysis, we observed that a clear, interpretable pattern is rare across layers and heads. This result is in accordance with visualizations provided in \cite{alphacode} \footnote{https://alphacode.deepmind.com/}, where a far stronger model exhibits patterns that can be not so intuitive. In Figures \ref{fig:before_1},\ref{fig:after_1},\ref{fig:before_2}, \ref{fig:after_2} we observe minor differences between non-finetuned and finetuned versions. 
The underlying changes in the reasoning processes of our coding models are not directly visible with attention maps. Reasoning processes should be viewed as an effect emergent from multiple interactions across layers and heads and can thus not always be located in a specific part of them.

\end{document}